\documentclass[letterpaper,twocolumn,10pt]{article}
\usepackage{usenix,epsfig,endnotes}
\usepackage{comment,soul,times}
\usepackage{graphicx,tabulary}
\usepackage{amsmath,amssymb}
\usepackage{hyperref,authblk}

\usepackage{color}
\usepackage{amsmath}
\usepackage{amssymb}
\usepackage{multirow, varwidth}
\usepackage{stfloats}
\usepackage{verbatim}
\makeatletter
\newif\if@restonecol
\makeatother

\usepackage[ruled,vlined,lined,boxed,linesnumbered]{algorithm2e}
\let\oldnl\nl% Store \nl in \oldnl
\newcommand{\nonl}{\renewcommand{\nl}{\let\nl\oldnl}}
\usepackage{xr}
\usepackage[amsmath,thmmarks]{ntheorem}

\theorembodyfont{\normalfont}
\theoremseparator{.}
\theoremstyle{nonumberplain}

\DeclareRobustCommand\onedot{\futurelet\@let@token\@onedot}
\def\onedot{. }
\def\eg{\emph{e.g}\onedot} 
\def\ie{\emph{i.e}\onedot}

\def\etal{\emph{et al}\onedot}

\newcolumntype{K}[1]{>{\centering\arraybackslash}p{#1}}
\usepackage{color}

\newcommand{\qirong}[1]{{\color{cyan}{\small\bf\sf [Qirong: #1]}}}

\begin{document}
%don't want date printed
\date{}

%make title bold and 14 pt font (Latex default is non-bold, 16 pt)
\title{Poseidon: A System Architecture for Efficient GPU-based Deep Learning on Multiple Machines}
\author{
Hao Zhang$^1$, Zhiting Hu$^1$, Jinliang Wei$^1$, Pengtao Xie$^1$\\
\vspace{-8pt}
       Gunhee Kim$^2$, Qirong Ho$^3$ and Eric P. Xing$^1$
       \\
       $^1$Carnegie Mellon University \ $^2$Seoul National University \ $^3$Institute for Infocomm Research\\
       \small \{hao, zhitingh, jinlianw, pengtaox, epxing\}@cs.cmu.edu, gunhee@snu.ac.kr, hoqirong@gmail.com\\
}

\maketitle

% Use the following at camera-ready time to suppress page numbers.
% Comment it out when you first submit the paper for review.
\thispagestyle{empty}

\subsection*{Abstract}
Deep learning models, which learn high-level feature representations from raw data, have become popular for machine learning and artificial intelligence tasks that involve images, audio, and other forms of complex data. A number of software ``frameworks'' have been developed to expedite the process of designing and training deep neural networks, such as Caffe~\cite{Jia:2014:MM}, Torch~\cite{Collobert:2011:NIPSW}, and Theano~\cite{Bergstra:2011:NIPSW}. Currently, these frameworks can harness multiple GPUs on the same machine, but are unable to use GPUs that are distributed across multiple machines; because even average-sized deep networks can take days to train on a single GPU when faced with 100s of GBs to TBs of data, distributed GPUs present a prime opportunity for scaling up deep learning. However, the limited inter-machine bandwidth available on commodity Ethernet networks presents a bottleneck to distributed GPU training, and prevents its trivial realization.

To investigate how existing software frameworks can be adapted to efficiently support distributed GPUs, we propose \textit{Poseidon}, a scalable system architecture for distributed inter-machine communication in existing deep learning frameworks. In order to assess Poseidon's effectiveness, we integrate Poseidon into the Caffe~\cite{Jia:2014:MM} framework and evaluate its performance at training convolutional neural networks for object recognition in images. %which builds upon the Petuum distributed ML framework~\cite{Xing:2015:KDD} as a starting point.
Poseidon features three key contributions that improve the training speed of deep neural networks on clusters: (i) a three-level hybrid architecture that allows Poseidon to support both CPU-only clusters as well as GPU-equipped clusters, (ii) a distributed wait-free backpropagation (DWBP) algorithm to improve GPU utilization and to balance communication, and (iii) a dedicated structure-aware communication protocol (SACP) to minimize communication overheads.
%to be applicable on any system configuration,
We empirically show that Poseidon converges to the same objective value as a single machine, and achieves state-of-the-art training speedup across multiple models and well-established datasets, using a commodity GPU cluster of 8 nodes
(\eg$4.5\times$ speedup on AlexNet, $4\times$ on GoogLeNet, $4\times $ on CIFAR-10).
On the much larger ImageNet 22K dataset, Poseidon with 8 nodes achieves better speedup and competitive accuracy to recent CPU-based distributed deep learning systems such as Adam \cite{Chilimbi:2014:OSDI} and Le \etal \cite{Le:2012:ICML}, which use 10s to 1000s of nodes.

\section{Introduction}

Deep learning (DL), which refers to a class of neural network models with deep architectures, forms an important and expressive family of machine learning (ML) models.
%Deep neural networks have been proved quite powerful, especially for learning high-level feature representations from raw data.
Modern deep learning models, such as convolutional neural networks (CNNs), have achieved notable successes in a wide spectrum of machine learning tasks, including speech recognition~\cite{Deng:2013:ICASSP}, visual recognition~\cite{Krizhevsky:2012:NIPS} and language understanding~\cite{Mikolov:2013:ICLRW}.
The explosive prosperity and rapid adoption of CNNs by research community are largely attributed to high performance computing hardware, such as GPUs, as well as a wide range of easy-to-use open source frameworks based on GPUs, including Caffe~\cite{Jia:2014:MM}, Torch~\cite{Collobert:2011:NIPSW}, Theano~\cite{Bergstra:2011:NIPSW}.
As of writing, the current, official versions of these toolkits can harness multiple GPUs on the same machine, but are unable to use GPUs that are distributed across multiple machines, which limits their practical use to smaller datasets.

%\qirong{You can keep this para in the intro, but you must be more careful of the tone. Mention that these systems demonstrate the potential of distributed deep learning, but have one flaw or another. Emphasize that our goal is to bring distributed training to an existing, well-established popular-but-single-machine platform, without commenting too much on other platforms.}
% justify the potential of scaling up deep learning using clusters
On the other hand, several CPU-based distributed systems for deep learning have been implemented. Zou \etal \cite{Zou:2014:VLDB} report the Tencent deep learning platform named as \textit{Mariana}, which distributes neural network training onto CPU clusters.
%leverages several useful features including parameter swapping, partitioned linearity topology, and approximate AdaGrad optimization.\qirong{No need to talk about ALL of Mariana's technical features. It will distract from your point. Just say what size and scale they achieve, and their most important single technical development.}
Google's \textit{DistBelief} framework \cite{Dean:2012:NIPS} allows training deep networks on CPU-only clusters with up to 1,000 machines, while Le \etal\cite{Le:2012:ICML} later scale up to a cluster of 16,000 CPU cores by exploiting model parallelism and asynchronous SGD.
%Wang \etal\cite{Wang:2015:MM} propose \textit{SINGA}, a distributed platform to support different deep learning models for multiple multimedia applications.
Recently, Microsoft's Adam \cite{Chilimbi:2014:OSDI} achieved state-of-the-art results on the ImageNet22K classification task, by leveraging distributed systems techniques such as a global parameter server, cache locality, and staleness control between workers.
These frameworks demonstrate that there is excellent potential to scale up deep learning using distributed clusters, though they require large clusters with thousands of CPU cores to produce the reported results.
%\haow{Though these frameworks may suffer one or two flaws, %including that are However, most of them are not openly available for use by the general research community, and furthermore
%such as that they are all built upon CPUs, thus requiring thousands of CPU cores to produce the advertised results,
%Such speedups could potentially be achieved with a smaller number of GPU-equipped machines, which are more readily available to researchers.
%}

%\hao{Argue for GPU based distributed deep learning, and rise the technical problems we need to handle}
Compared to CPU-based distributed deep learning, parallelization of deep networks on GPU-equipped clusters is more readily available to researchers, since satisfactory speedups could potentially be achieved with a smaller number of GPU cards \cite{Krizhevsky:2014:arXiv}.
However, different from the setting of a single machine with multiple GPUs where near-linear speedups could be trivially realized, scaling up deep learning on multiple GPU-equipped machines faces two major challenges.
First, Infiniband networking, which has been responsible for past successes in distributed DL \cite{Coates:2013:ICML}, is not available on most cloud computing platforms and lab clusters, where only commodity hardware with limited network bandwidth is installed.
Since GPUs are often orders-of-magnitude faster in matrix-dense computations compared to CPUs, in GPU-based distributed training, gigabytes of parameters are generated per second on each device, waiting to be synchronized across multiple machines. Such a high communication load raises the network communication as the main bottleneck given limited bandwidth of commodity Ethernet. Second, managing the computation and communication in a distributed GPU cluster often complicates the algorithm design.
Consequently, more algorithm-specific strategies and dedicated communication protocols are necessary to attain maximum performance when designing GPU-based distributed DL.

%\qirong{our goal is to enhance existing popular single-machine platforms with distributed GPU capability, not to build a new framework. Please change, following the new abstract.}
%\hao{Argue that transforming an existing framework has merit}
In this paper we investigate how existing software frameworks can be adapted to efficiently support distributed GPUs, given that only commodity Ethernet is available. On one hand, instead of building a new DL framework from scratch, our goal is to develop an efficient system engine for distributed deep learning, and thus enhance existing popular single-machine platforms with distributed GPU capability. Transforming an existing framework rather than designing a completely new one has the following merits: First, it preserves the ecosystem better and saves users the effort of making an expensive switch. Second, it enable us to solely focus on designing fast and efficient distributing strategies, at the same time enjoy any algorithmic advantage brought by the third-party DL framework themselves. On the other hand, in contrast to systems that require specialized hardware~\cite{Coates:2013:ICML}, we want our solution to effectively harness distributed GPUs installed on commodity servers and connected via Ethernet, so that our software is as accessible as possible to researchers.
To this end, we propose an open-source system architecture, Poseidon\footnote{Poseidon was initially released in January 2015 along with Petuum v1.0 as an application under the B\"{o}sen parameter server, with GPU support added in July 2015. All source codes are available at \url{github.com/petuum/poseidon/}.}, which can be deployed on a variety of cluster configurations (such as CPU-only clusters, or GPU-equipped clusters, or clusters with multiple GPUs per machine).
Poseidon makes use of any existing single-machine DL framework, and implements a distributed system layer underneath it, in order to harness distributed CPU and GPU clusters with commodity hardware. In our current implementation, we chose Caffe because of its popularity, while noting that Poseidon's design is compatible with other CNN libraries such as Torch and Theano.

%\qirong{Move the Petuum reference to the end. Start first by talking about your unique contributions, and then mention at the end that we built our system by adapting Petuum.}

In order to efficiently distribute DL on GPU clusters, we propose three key contributions:
First, we design Poseidon as a hybrid three-level architecture, which allows Poseidon to work on both CPU-only as well as GPU-equipped clusters. Second, we propose \textit{distributed wait-free backpropagation} (DWBP), which leverages the chain rule in backpropagation (BP) and the structure of modern CNNs; DWBP improves GPU utilization and balances communication load, by overlapping computation with communication during BP. Third, we develop a \textit{structure-aware communication protocol} (SACP), which combines a centralized parameter storage with decentralized peer-to-peer broadcasting, to minimize communication overheads.
Together, these three components allow Poseidon to address the communication bottleneck in GPU-based DL on commodity clusters --- specifically, how to efficiently synchronize parameters across Ethernet networks, particularly when each GPU can generate Gbs of gradients per second. We implemented Poseidon's distributed layer upon the Petuum distributed ML framework~\cite{Xing:2015:KDD}, which provides a bounded stale synchronous parallel (SSP) parameter server~\cite{Ho:2013:NIPS} that preserves data-parallel convergence guarantees, and prioritized network bandwidth allocation~\cite{Wei:2015:SoCC}.

%\qirong{Please tie these results to the fact that Poseidon is a solution to the network bottleneck.}
Poseidon significantly reduces the training time required by state-of-the-art CNN models, while still achieving the same quality of convergence and accuracy. Using a cluster of 8 GPU-equipped Ethernet-connected commodity machines, by significantly alleviating the bottleneck issue raised by the limited bandwidth, Poseidon attains almost the same classification accuracy as a single GPU, but is roughly $4.5\times$ faster when training AlexNet, and $4\times$ faster when training GoogLeNet. These results hold across benchmark datasets of different sizes: CIFAR-10~\cite{Krizhevsky:2009:cifar}, ILSVRC2012, and ImageNet 22K~\cite{Russakovsky:2015:IJCV}. For example, on a small task such as CIFAR-10 quick solver (where distributed training might not be expected to perform well), 8-node Poseidon can achieve better accuracy than a single machine, in 1/4-th the time. To demonstrate the scalability of Poseidon, we train CNN classifiers on the ImageNet22K dataset, consisting of 14.2M images in 21,841 categories, and achieve competitive accuracy with state-of-the-art results, in less training time and using fewer machines (\eg 30\% training time and 13\% cluster nodes compared to Adam \cite{Chilimbi:2014:OSDI}). %

%\qirong{please change the contributions to follow the emphasis in the abstract. e.g. (for this sytem paper only) Posiedon is not a framework, but a system to enable single-machine frameworks to go distributed.}
We summarize our main contributions as follows: (1) We propose \textit{Poseidon}, a scalable system architecture as a general purpose solution for any single-machine DL framework to be efficiently distributed on GPU clusters with commodity Ethernet, by leveraging the Petuum framework~\cite{Xing:2015:KDD} as well as three components: a three-level architecture, distributed wait-free backpropagation, and structure-aware communication protocol.
(2) We empirically show that Poseidon, running on a GPU-equipped cluster with commodity hardware and Ethernet, achieves high quality convergence comparable to a single machine, as well as state-of-the-art training speedups on benchmark CNN classification models (\eg $4.5\times$ on AlexNet, $4\times$ on GoogLeNet, $4\times$ on CIFAR-10, over a single machine) --- even for larger datasets such as ImageNet 22K, Poseidon achieve competitive accuracy as compared to the state-of-the-art results, but using only $30\% $ training time and $13\%$ cluster nodes.

The rest of the paper is organized as follows. In section \ref{sec:relatedwork}, we review existing works on GPU-based distributed DL. Section \ref{sec:background} covers the basics of neural network models, and briefly introduces some fundamentals of Petuum PS and data-parallel distributed machine learning. In section \ref{sec:poseidon}, we present the architecture and key features of Poseidon. Section \ref{sec:evaluation} evaluates Poseidon on multiple standard dataset with regard to efficiency, scalibility and accuracy. Section \ref{sec:conclusion} concludes the paper.

\section{Related Work}
\label{sec:relatedwork}

%\qirong{This paragraph goes to related work. Your discussion is focused on general software frameworks, not instances of parallel deep learning.}
Because of the demand for faster training of neural networks on ever-larger datasets, several frameworks have been proposed that use multiple GPUs on a single machine. For example, Yadan \etal \cite{Yadan:2014:ICLRW} show that mixed parallelism yields better speedups over model-only or data-only parallelism in ImageNet classification with 4 GPUs. Similarly, Krizhevsky~\cite{Krizhevsky:2014:arXiv} also implements mixed parallelism for AlexNet~\cite{Krizhevsky:2012:NIPS} with 8 GPUs which relies on data parallelism in the convolutional layers and on model parallelism in the fully-connected layers. Facebook's \textit{fbcunn}~\cite{fbcunn,Vasilache:2015:ICLR} implements both model- and data- parallelism on multiple GPUs.
However, the aforementioned frameworks focus on parallelization within a single machine with multiple GPUs, and cannot take advantage of distributed computing environments where GPUs are spread out across a cluster.

Distributed, multi-node GPU-based CNN training is an active area of research.
Coates \etal\cite{Coates:2013:ICML} demonstrated that they could train a 11-billion parameter network on a cluster of 16 GPU nodes using model-parallelism, but their implementation required specialized hardware, such as Infiniband networking. \textit{MXNet}
%~\cite{MXNet}
is an open-source framework for distributed deep learning,
%that integrates key features from several other deep learning systems: \textit{Minerva}~\cite{Wang:2014:NIPSW}, \textit{Purine}~\cite{Lin:2015:ICLR} and the parameter server of~\cite{Li:2014:NIPS}.
that addresses both algorithmic code for DL, which is the role that Caffe plays in this paper, as well as distributed execution, which is the technical focus of this paper. No peer-reviewed results for MxNet are available as of writing.
%this combined approach has reported promising results\footnote{\url{learningsys.org/papers/LearningSys_2015_paper_1.pdf}.}.

Our position is to identify reusable systems techniques that can be applied to {\it existing} single-machine DL frameworks in order to add value to their mature userbase and software ecosystem. We choose Caffe as our example, but note that our techniques could be used for other single-machine deep learning software such as Torch, Theano. Moreover, we make use of commodity hardware (\eg machines with 1-2 GPUs and Ethernet networking) instead of specialized hardware that is not readily available from cloud providers or most academic clusters (\eg Infiniband or machines with $\ge 4$ GPUs). Through our work, we hope to enable existing popular frameworks to be scaled up to distributed clusters of GPU machines.

Recently, Google released their TensorFlow software for deep learning,
%\footnote{\url{tensorflow.org}.},
which does not currently support distributed GPU training, and does not have peer-reviewed results. As with Caffe, we believe the techniques presented herein could be used to produce a distributed version of Tensorflow. Also of note are several efforts to port Caffe onto the Spark platform, such as SparkNet~\cite{moritz2015sparknet}, which reports a 4-5 times speedup with 10 machines (and hence less scalability than our results herein), as well as a recent, non-peer-reviewed, effort by Yahoo
%\footnote{\url{yahoohadoop.tumblr.com/post/129872361846/large-scale-distributed-deep-learning-on-hadoop}.}
which exclusively uses Infiniband RDMA. In contrast, our focus is on {\it commodity Ethernet} that is readily available in most clusters and cloud providers. We see SparkNet in particular as closest to the spirit and intent of this paper; namely, to scale up existing deep learning frameworks with generic, re-usable distributed techniques, and thus add value to their mature ecosystems.

%\hao{Need more careful arguments}
%MXNet is still under active development, and has recently reported results on ImageNet22K classification using a single machine with 4 GPUs. However, MXNet's performance when using a cluster of multiple GPU-equipped nodes has not been reported yet; we evaluate Poseidon on a cluster of 8 GPU-equipped nodes.
%Google recently release the TensorFlow \hao{Need argument}

%However, the current release of TensorFlow only supports single machine training, and it performance is not yet evaluated using standard tasks and datasets.

%\hao{Do we need to include a branch of newly released GPU frameworks? like Yahoo, IBM, etc}
% and thus its performance on distributed GPU clusters has not been reported yet with any standard datasets, in terms of throughput, convergence acceleration ratio, and accuracies.
%Moreover, it is not equipped yet with dedicated communication strategies to boost performance with GPU clusters.
%\qirong{I have removed statements which cannot really stand as criticisms, such as ``only strict model parallelism'', or ``no dedicated communication strategies''. In general, don't criticize system designs unless we have very strong evidence. Let our results speak for themselves.}

\section{Background}
\label{sec:background}
%\qirong{It is not good style to have 2 section headings in a row. Begin with some intro/motivation or a high-level overview of what you are going to talk about, and why.}
Poseidon builds upon an existing general-purpose system for distributed machine learning algorithms, Petuum, and extends it with new contributions that specifically improve the performance of GPU-based deep learning. In order to clearly delineate our contributions, we begin with a brief overview of the Petuum features that we build upon, and establish some mathematical notations that will be useful in characterizing Poseidon.

\subsection{Petuum for Iterative-Convergent ML}
\label{sec:Petuum}
Poseidon builds upon Petuum, a distributed big machine learning framework that provides a generic interface to a broad spectrum of ML programs~\cite{Xing:2015:KDD}. Its design philosophy is rooted in \textit{iterative-convergent} solutions to loss function minimization. A number of ML algorithms are formulated in this manner, which involves repeatedly executing update equations that decrease some error functions. Some notable examples include stochastic gradient descent in optimization programs, MCMC and variational methods for graphical models, and proximal optimization for structured sparsity problems, among others.

In a mathematical form, the iterative-convergent algorithm can be represented as follows. Given data $D$ and a loss function $\ell$, a typical ML problem can be solved by iteratively executing the update equation until the model parameters $A$ reaches some stopping criteria.
\begin{equation}
\label{eq:iter_conv}
A^{(t)} =  F (A^{(t-1)}, \Delta_{\ell} (A^{(t-1)}, D))
\end{equation}
\noindent where $t$ denotes the iteration.
The update function $\Delta_{\ell}$ performs computation on data $D$ with model parameters $A$ to improve the loss $\ell$. The intermediate results are aggregated by function $F$.

\begin{figure}[t]
%\hspace{-1.5em}
\includegraphics[width=80mm]{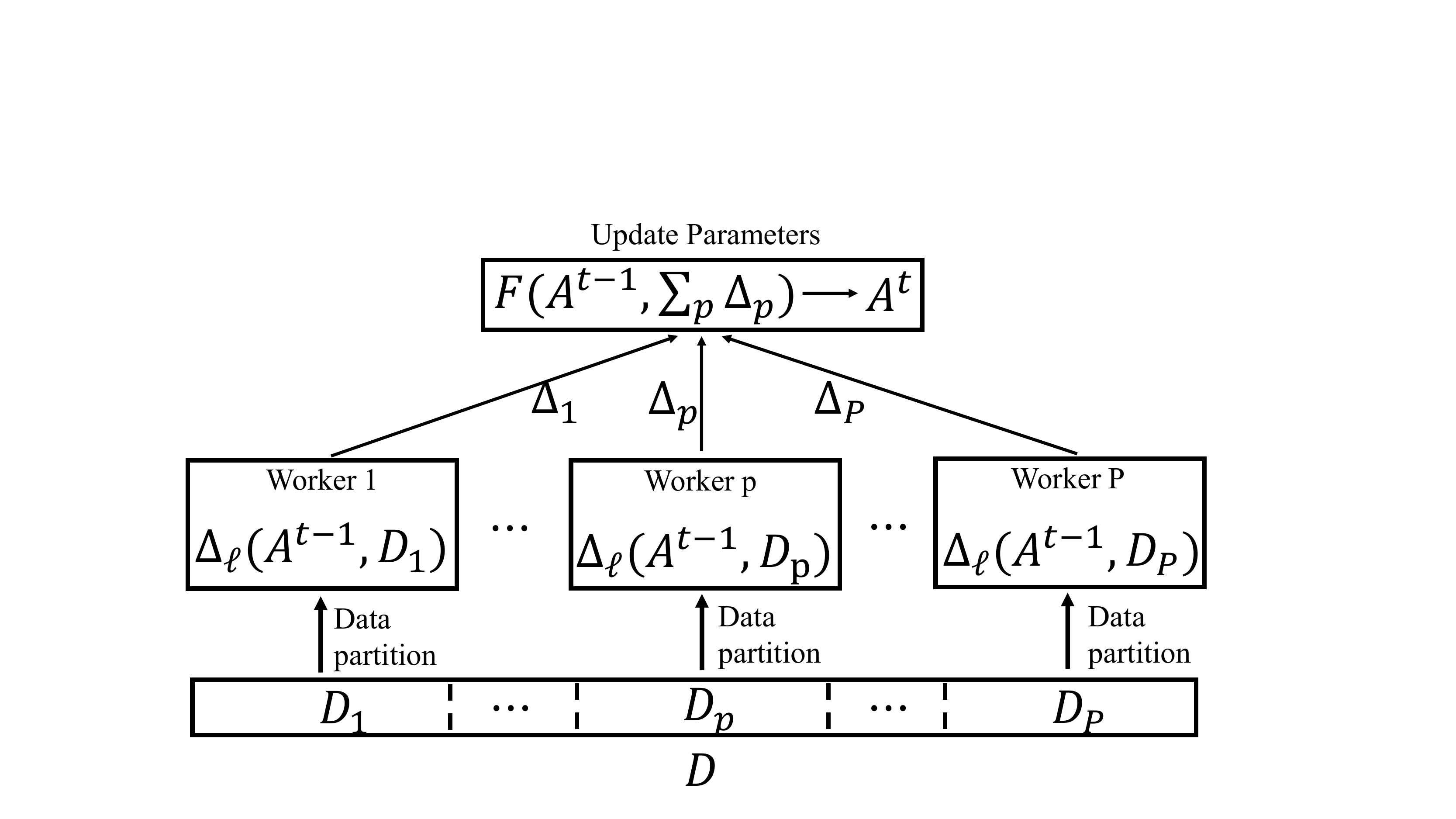}
\caption{
The iterative-convergent algorithm in a data-parallel distributed setting.}
\label{fig:data_parallel}
\end{figure}

In large-scale machine learning, both data $D$ and model $A$ can be very large. In data-parallelism, the data $D$ is partitioned and assigned to computational worker machines (indexed by $p=1,\cdots,P$), whereas in model-parallelism, the model $A$ is partitioned and assigned to workers. Since we are interested in data-parallelism, we partition the data $D$ into a set of $D_p$ denoting the $p$-th data partition (\ie often called mini-batch), as shown in Figure \ref{fig:data_parallel}. Then, the update equation becomes
\begin{equation}
\label{eq:iter_conv_data_paral}
A^{(t)} =  F (A^{(t-1)}, \sum_{p=1}^P \Delta_{\ell} (A^{(t-1)}, D_p))
\end{equation}
In each iteration, parameter updates $\Delta_{\ell}$ produced by each partition of data are locally computed on each worker, and then are communicated to each other.

\subsection{Stale Synchronous Parallel PS}
\label{sec:ps}
A parameter server (PS) is a distributed shared memory system that provides systematic abstraction for iterative-convergent algorithms in data-parallel distributed machine learning. Typically, PS enables each worker to access the global model parameters $A$ via network communications following the client-server scheme. In particular, the training data are partitioned and distributed to a large number of clients (\ie workers). Data-parallel distributed training can be easily implemented on the PS architecture, by letting the execution of the update $\Delta(\cdot)$ take place only on each worker over data subsets therein, and the application of the updates to model parameters $A$ take place on the server, and a consistency scheme coordinate the synchronization among server and clients.

In data-parallel ML, iterative-convergent algorithms often enjoy a nice property of error-tolerance, \ie they still execute and converge correctly even when their model parameters $A$ experience synchronization delays, provided that those delays are strictly bounded ~\cite{Ho:2013:NIPS,Dai:2015:Analysis,Kumar:2014:Fugue}. The stale synchronous parallel (SSP) consistency model exploits this error-tolerance property, and try to reduce network communication/synchronization overheads substantially by allowing stale parameter updates while the staleness is bounded by a threshold $s$.
Integrated with a PS, the SSP consistency model ensures that if a worker reads from server at iteration $t$, it is guaranteed to receive all updates from all workers computed at and before iteration $t-s-1$. If this is impossible because some straggling worker is more than $s$ iterations behind, the reader will stop until the straggler catches up and sends its updates. For stochastic gradient descent algorithms, SSP has very attractive theoretical properties \cite{Dai:2015:Analysis}.

Poseidon's distributed layer is derived from {\it B\"{o}sen}~\cite{Wei:2015:SoCC}, a parameter server implementation that supports SSP consistency model. It allows computations to use stale model parameters (to reduce synchronization overheads), but strictly upper-bounds the number of missing iterations, restoring formal convergence guarantees~\cite{Ho:2013:NIPS}.　Besides B\"{o}sen and SSP, Poseidon provides many advanced features that are beneficial for GPU-based distributed deep learning, as covered in section \ref{sec:other}.

\subsection{Data-parallel Distributed Training of Convolutional Neural Networks}
\label{sec:cnn}
A neural network has multiple stacked layers, each of which is filled with different types of computing units inside, and layer-wisely interconnected by real or boolean weight matrices as trainable parameters.
The basic computing unit in each layer is called a neuron, which is usually composed of a vector of weights corresponding to a row in the weight matrix, and a nonlinear function to introduce rich model expressiveness. Each neuron takes outputs (activations) from its preceding layer as input, applies both linear and nonlinear transformations to produce its own activation, which is then passed to its following layers as their input. At the bottom of a neural network is an input layer reading and vectorizing different types of data as network inputs, while at the top of the network is usually a loss layer, which are pre-specified by an optimization objective (\eg a classifier or a regressor).

Convolutional neural networks (CNNs) have both convolutional layers and fully-connected layers as building blocks. A neuron in a convolutional layer is also called a filter, and is connected with a spatial local region of its previous layer's output (feature maps), and share the same weights across all possible regions. This weight sharing pattern significantly reduces the number of trainable parameters, making them much easier to train and more agnostic to overfitting. A convolutional layer is usually followed by a nonlinear down-sampling layer, such as an max-pooling layer, which partitions the output feature map into a set of rectangles and outputs the maximum value for each such sub-region.

CNNs are trained using the stochastic gradient descent (SGD), which falls into the family of iterative-convergent algorithms. Specifically, training is performed by an iterative algorithm, where each iteration consists of a feedforward and a backpropagation pass. In the feedforward pass, the network takes a batch of training samples as input, forwards from bottom to top layers and outputs a prediction for each sample at the end layer. The predictions are then compared to the groundtruth of training samples at the loss layer to compute the error value. In the backpropagation, the error is propagated through the network in a reverse order, during which the weights in each layer are updated towards the direction where the loss decreases. After repeating a sufficient number of training iterations, the network will usually converge to some state where the loss is close to an optimal, and the training is then terminated.

Accordingly, learning CNNs is another typical distributed ML problem to which the Petuum's iterative-convergent strategy is successfully applicable. In the CNN training, the update equation Eq.\ref{eq:iter_conv} reduces to
\begin{equation}
\label{eq:iter_conv_cnn}
A^{(t)} =  A^{(t-1)} +  \epsilon \cdot \nabla_{\ell} (A^{(t-1)}, D_p) + \Lambda(A^{t-1})
\end{equation}
where the parameter updates are calculated as the gradients of $A$ over current data batch $D_p$, controlled by a stepsize $\epsilon$, and the updating function $F$ reduces to the additive function as in SGD. We often impose a function $\Lambda(A^{t-1})$, which contains regualization and momentums on the model parameters $A$.

Similarly, in the data-parallel distributed setting, every node holds a replica of the network parameters $A$. At each iteration, every node $p$ takes a batch of data $D_p$, performs a feed forward and back propagation pass, and produces a copy of gradients. Gradients are then communicated, aggregated, and applied to update model parameters as
\begin{equation}
\label{eq:iter_conv_data_parallel_cnn}
A^{(t)} \hspace{-1pt}=\hspace{-1pt}  A^{(t-1)} \hspace{-2pt}+  \sum_{p=1}^{P} \left[ \epsilon \hspace{-1pt}\cdot \hspace{-1pt} \nabla_{\ell} (A^{(t-1)}\hspace{-1pt}, D_p) \hspace{-1pt}+\hspace{-1pt} \Lambda(A^{t-1}) \right].
\end{equation}

\section{Poseidon Architecture}
\label{sec:poseidon}

\begin{table}[tbp]
\footnotesize
\centering
	\begin{tabular}{|c|c|c|c|}
	\hline
	Ethernet & Rate(GBit/s) & Rate (Mb/s)  &  Rate (\# floats/s) \\ \hline \hline
	{\textbf{1 GbE} } & 1 & 125 & 31.25M     \\ \hline
	{\textbf{10 GbE} } & 10 & 1250 & 312.5M  \\ \hline
  	{\textbf{Infiband} } & 40 & 5000 & 1250M  \\ \hline
    \end{tabular}
\caption{The maximum throughput that commonly used Ethernet can provide in terms of how many Gigabits, Megabytes and number of float parameters could be transferred per second.}
\label{tb:Ethernet}
\end{table}

\begin{table}[tbp]
\footnotesize
\centering
    \begin{tabular}{|c|K{1.2cm}|K{1.5cm}|K{0.7cm}|K{1.2cm}|}
	\hline
	Model & Batch size (\# images) &\# parameters (\# floats) & Time (s/iter) & Gradients (\# floats/s) \\ \hline \hline
	{\textbf{AlexNet} }  & 256 & 61.3M  & 0.96 &  63.85M \\ \hline
	{\textbf{VGG-16} } & 64 & 128.3M  & 4.06 &  31.60M\\ \hline
    \end{tabular}
\caption{Statistics of modern CNN training, including the batch size, number of model parameters, per-iteration computation time and number of gradients generated per second on a single device. The performance is evaluated on a K40 GPU with standard solver settings, as reported in the official site of Caffe.}
\label{tb:cnn_time}
\end{table}

Because GPUs are faster in matrix computations than CPUs, the gradient updates are produced faster on GPUs than they can be naively synchronized over the network, thereby the computations during neural network training are usually bottlenecked by communications, as evidenced by Table \ref{tb:Ethernet} and Table \ref{tb:cnn_time}. In particular, Table \ref{tb:Ethernet} lists the standards of commonly used Ethernet and Table \ref{tb:cnn_time} shows some statistics of modern CNN training \footnote{The performance is quoted from the official site of Caffe: \url{caffe.berkeleyvision.org/performance_hardware.html} and  \url{github.com/BVLC/caffe/issues/1317}.}. Take the AlexNet training as an example: Given a standard solver setting with batch size 256, 61.3 million of gradients will be generated per second on each device. If we distribute the training onto a commodity cluster with 8 nodes each equipped with 1 GPU, ideally the master node need to receive at least 490M float parameters, and then send out another 490M in one second to guarantee that the next iteration of computation on workers will not be blocked. Though adjusting the network configurations (\eg increasing batch size) may decrease the communication load, the demanded throughput is still far above the maximum throughput that the commodity Ethernet (\ie 1 GbE and 10GbE Ethernet) can afford\footnote{Also note that due to issues related with network protocols and software implementations, the actual performance we could achieve in practice is usually lower than standard values as reported.}. Therefore, when distributing DL on GPU clusters, the major challenges are how to quickly collect and aggregate the gradients, and how to efficiently synchronize updated parameters across all workers.

%\qirong{Hao, you need to provide evidence and explanation as to WHY these are challenges. We know it's because GPUs produce updates so quickly relative to network bandwidth --- please give citations or numbers or our own results, etc..}

Poseidon presents three key contributions to address these challenges: a three-level hybrid architecture that supports both CPU and GPU computation, a distributed wait-free backpropagation (DWBP) scheme to interleave computation with inter-machine communication, and a structure-aware communication protocol (SACP) that reduces the size of network messages. The three-level architecture improves Poseidon's generality, by allowing it to work with both CPU- and GPU-based DL software, while DWBP and SACP enable the DL software to communicate quickly and efficiently across the network.

%\qirong{In general, the following subsections talk about *how* we do things, but don't say *why* we need to do them. You need to remind readers with evidence and justification as to why we have to go to such trouble. Why can't we just naively send the gradients across the network? What happens when you do that? Include evidence or refer to our results section to show why.}

\subsection{Overview: A Three-level Structure}
\label{sec:overview}

\begin{algorithm}[t]
\label{algo:overview}
\caption{CNN training with data-parallelism on Poseidon}
\small
\nonl \hspace{-1em}\textbf{Slave nodes:}\\
\setcounter{AlgoLine}{0}
 Partition training data $D$ equally into $\{ D_i \}_{i=1}^P$ and distribute them to all $P$ nodes.\\
 Replicate the initial model parameters $A$ to every worker thread $p$ as $A_p$.\\
 \For{$t = 1$ \KwTo $T$}{
	\ForEach{ worker thread $p \in \{1, 2, \cdots, P\}$}{
	Take a batch of training data $D_p^t$ from $D_p$. \\
	Perform forward pass.\\
	Perform backpropagation pass following the DWBP algorithm (See Algorithm.\ref{algo:dwbp}).\\	
	Update local synchronization states to the consistency manager (see section \ref{sec:ps}).
}
}
\setcounter{AlgoLine}{0}
\nonl \hspace{-1em} \textbf{Master node:}\\
 \For{$t = 1$ \KwTo $T$}{
 Collect gradients that are sent by worker nodes. \\
 Updates the part of model parameters for which a corresponding gradient is received. \\
 Push updated model parameters to worker nodes according to the consistency manager.\\
}
\end{algorithm}

\begin{figure}[t]
\includegraphics[width=80mm]{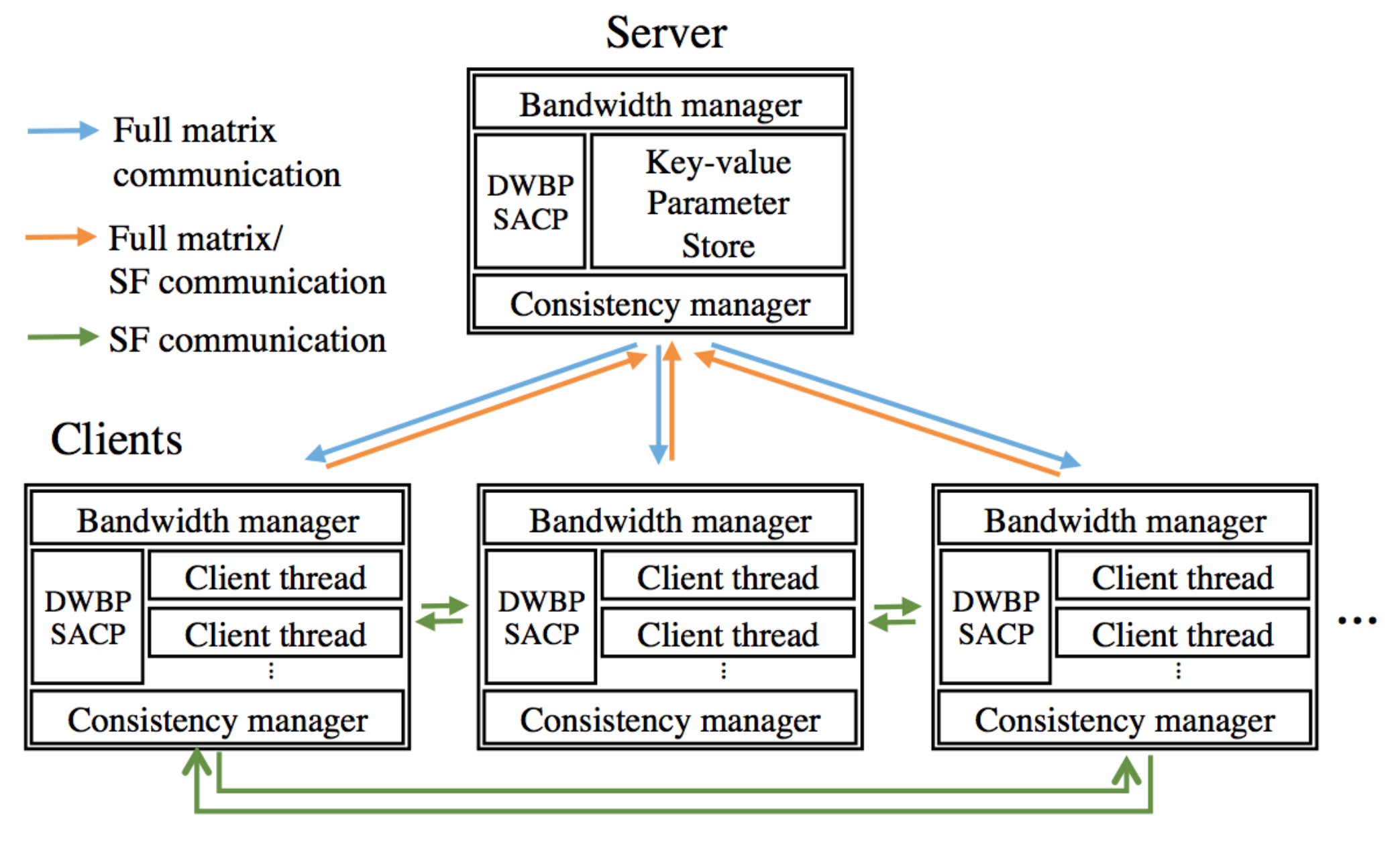}
\caption{
An overview of the distributed architecture of Poseidon.}
\label{fig:overview}
\end{figure}

Existing systems for distributed deep learning usually exhibit a traditional client-server structure. For example, in previous CPU-based distributed DL systems \cite{Chilimbi:2014:OSDI, Dean:2012:NIPS}, a two-level parameter server architecture was built, where the first level has server machines collecting gradients and distributing newly updated model parameters to workers, and the second level has worker nodes (threads) taking batches of training data and generating gradient updates. When deploying them onto GPU clusters, one may need to heavily adjust the implementation, to support more sophisticated cluster configurations ( \eg a cluster of GPU nodes where each node has multiple GPUs), as well as to avoid unnecessary memory access between different types of devices. Moreover, existing architectures only allow connections between server and clients, which limits that the communication can only happen between master and slave nodes.

In order to provide a general solution for both CPU-only and GPU-based distributed deep learning as well as to enable more strategic communication approaches, we design Poseidon as a three-level structure, as Fig.\ref{fig:overview} illustrates.
First, We add an additional hierarchy within each worker node, thus allow multiple client threads coexisting in a single worker machine. This design enables Poseidon to support both CPU and GPU users as well as any system configuration, such as a cluster of nodes where each node has multiple GPUs or CPU cores, by binding each worker thread with a specific device (CPU core or GPU).
%\pengtao{Multi-threads within each worker machine has been implemented in Yahoo LDA and Petuum PS. This is not something new.}
%Second, Poseidon creates different thread caches for different client threads in the memory of each worker, caching elements of the model parameters $A$.
%\jinliang{"different libraries"?? -- I think you meant "thread cache"?} % for faster access and consistency management.
%It avoids unnecessary memory access between different computational devices by leveraging the fact that the memory exchange in the same type of devices (\eg between GPUs) is usually costless. Without the thread caches, the
Second and more importantly, instead of the traditional client-server structure, where each client only connects with the server machine, we design a hybrid topology, where peer-to-peer (P2P) connections between pairs of workers, and server-client connections between the server and workers, are both established. It enables more dedicated communication strategies for parameter synchronization among multiple-GPU nodes, which we elaborate in section \ref{sec:samp}.

Algorithm \ref{algo:overview} presents an overview of the distributed training process of Poseidon. At the beginning of training, every worker thread starts its Caffe engine~\cite{Jia:2014:MM} to perform feedforward and then backpropagation pass for some number of times, via the distributed wait-free backpropagation (DWBP) algorithm (See section \ref{sec:dwbp}), during which they communicate asynchronously following a consistency model of the bounded stale synchronous scheme~\cite{Ho:2013:NIPS}, as we briefly introduced in section \label{sec:ps}. The DWBP algorithm enables communication to be overlapped with the error propagation computations. The structure-aware communication protocol (SACP) minimizes communication load by exploiting the layer property of neural nets, and  passing or receiving the parameter updates by intelligently choosing the optimal solution from the client-server or P2P pipelines (See section \ref{sec:samp}). In the lower level, the communications are further monitored and operated by a bandwidth manager provided by Petuum \textit{B\"{o}sen} \cite{Wei:2015:SoCC}, as we explain in section \ref{sec:bandwidth}.

\subsection{Distributed Wait-free Backpropagation}
\label{sec:dwbp}

%\qirong{What happens when you naively send gradients at the end of each iteration? Why must we interleave them this way?}
%\hao{I think I have elaborated the reasons with sufficient technical details in this section, and also give some illustrations in Figure \ref{fig:dwbp}. I also gave internal comparisons in section \ref{sec:internal_dwbp_sacp}. I will add a pointer for the readers to refer to the internal comparison results.}
%\qirong{Again, you need to bring this up front. Challenges first, then our solution, and then (if the arguments are available) why others don't work. Think about how you started section 4.0, with the challenges.}
Backpropagation (BP) \cite{rumelhart1985learning} is  the principle algorithm for training neural networks. Specifically, BP algorithm proceeds as a chain, with many feedforward and backpropagation passes. During the back pass, an error message $E$ is propagated from the top to the bottom of the network, thus a message passing chain is formed.

\begin{figure}[t]
\centering
\includegraphics[width=70mm]{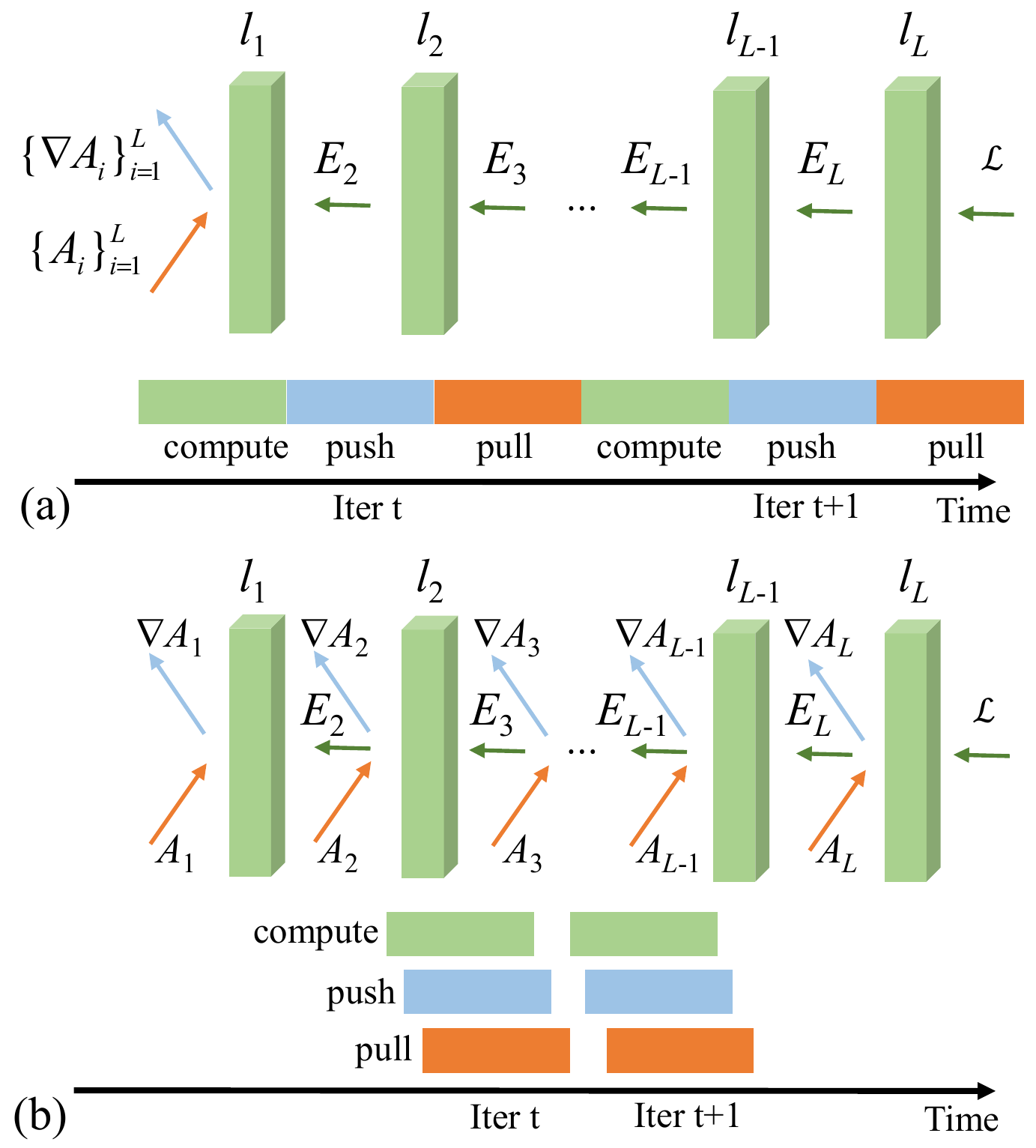}
\caption{\small
Comparison between (a) traditional backpropagation and (b) distributed wait-free backpropagation.
}
\label{fig:dwbp}
\end{figure}

Figure.\ref{fig:dwbp}.(a) shows the process of the original BP in distributed settings on a neural net with $L$ layers $\{l_i\}_{i=1}^L$ and layer parameters as $A = \{A_i\}_{i=1}^L$. At each iteration $t$, every worker $p \in \{1, \cdots, P\}$ performs the BP computation separately. Only when the propagation reaches at the bottom layer $l_1$ (\ie all gradients $\nabla A^p = \{\nabla A_l^p\}_{i=1}^L$ are generated), each worker is ready to start communication. The worker sends out local parameter updates $\nabla A^p$, waits for the remote master node to collect, aggregate and apply the parameter updates $\{A^p\}_{p=1}^P$ from all workers, and then synchronizes with the master node via the network to fetch a new copy of updated parameters $A$ for next iteration $t+1$. Therefore, each worker cannot proceed to iteration $t+1$ until it receives all updated layer parameters $\{A_l\}_{i=1}^L$; the computation and communication occur sequentially as shown in Figure.\ref{fig:dwbp}.(a).

\begin{algorithm}[b]
\label{algo:dwbp}
\caption{The Distributed Wait-free Backpropagation (DWBP) Algorithm}
\small
\SetKwInOut{Input}{Input}
\nonl \hspace{-1em}\textbf{At iteration $t$ on worker $p$:}\\
\setcounter{AlgoLine}{0}
\nonl \hspace{-1em}\Input{Loss $\ell$.}
 \For{$i = L$ \KwTo $1$}{
    \If{$i == L$}{
        Compute gradients $\nabla A_{i} = \frac{\partial \ell}{\partial A_{i}}$ using $\ell$.\\
    }\Else{
        Receive error message $E_{i+1}$ from layer $l_{i+1}$.\\
        Compute gradients $\nabla A_{i} = \frac{\partial \ell}{\partial A_{i}}$ using $E_{i+1}$.\\
    }
    \If {$i \ne 1$}{
    Compute error message $E_{i}$  and pass to layer $l_{i-1}$.\\
    }
    Communicate: push out $\nabla A_{i}$ and pull in updated $A_i$ following the SACP protocol (See Algorithm \ref{algo:sacp});\\
}
\end{algorithm}

The distributed wait-free backpropagation is designed to reduce the waiting time of parameter synchronizations when backpropagation concurrently executes on multiple machines, so as to improve the GPU utilization. Specifically, leveraging the chain structure of BP, once layer $l_{i+1}$ finishes computations and propagates its error message $E_{i+1}$ to the preceding layer $l_i$, its gradients $\nabla A_{i+1}$ are ready to be sent out, and its parameters $A_{i+1}$ are also ready to be updated. This is because each layer $l_i$ in the network occupies an independent set of parameters $A_i$, and the subsequent computations of lower layers $\{l_1, \cdots, l_i\}$ do not affect upper layers $\{l_{i+1}, \cdots, l_L\}$ any more. Correspondingly, the parameter updating at upper layers $\{l_{i+1}, \cdots, l_L\}$ does not affect that of lower layers either, because the computations of layer $l_{i}$ only depend on the error message $E_{i+1}$, which have already been passed.

\begin{table}[tbp]
\footnotesize
\centering
	\begin{tabular}{|c|c|c|}
	\hline
	Parameters & CONV Layers (\#/\% ) & FC Layers (\#/\% )\\ \hline \hline
	{\tt AlexNet } & 2.3M / 3.75  & 59M / 96.25 \\ \hline
	{\tt VGG-16 } & 7.15M / 5.58 & 121.1M / 94.42 \\ \hline % 7152700 vs 121100000
    \end{tabular}

    \begin{tabular}{|c|c|c|}
	\hline
	FLOPs & CONV Layers (\#/\% ) & FC Layers (\#/\% )\\ \hline \hline
	{\tt AlexNet } & 1,352M / 92.0   & 117M / 8.0 \\ \hline
	{\tt VGG-16 } & 10,937M /  91.3 & 121.1M / 8.7 \\ \hline
    \end{tabular}
\caption{\small Parameter and FLOP distributions of convolution and fully-connected layers in AlexNet~\cite{Krizhevsky:2012:NIPS} and VGG-16~\cite{Simonyan:2015:ICLR}.}
\label{tb:distribution}
\end{table}

Algorithm \ref{algo:dwbp} with illustration of Fig.\ref{fig:dwbp}.(b) summarizes the DWBP algorithm, whose intuition is to concurrently schedule the computations of lower layers and the communications of upper layers during BP.
It exploits the chain structure of the network, and overlaps the communications at upper layers, with the computations at the lower layers. % (See Figure.\ref{fig:dwbp}(b)).
Different from the original BP, the DWBP enforces each layer to start its communication once its gradients are generated, and allows partial parameter updating on the layer.
Ideally, when the propagation reaches at the top of the network, both communication and computation are finished, thus the worker can immediately start next iteration.

% Why this is quite effective in modern CNN structure
The DWBP is even more effective in GPU clusters with state-of-the-art CNN architectures, such as AlexNet \cite{Krizhevsky:2012:NIPS} and VGG-16 \cite{Simonyan:2015:ICLR}, which stack convolutional (CONV) layers at the bottom, followed by fully-connected (FC) layers at the top. Table \ref{tb:distribution} shows the statistics about the sizes of parameters and computations in FLOPs for CONV layers and FC layers in AlexNet and VGG-16.
FC layers usually occupy more than $90\%$ of the model parameters, indicating communication costs are mostly consumed at the top FC layers,
while the CONV layers only take less than $10\%$ of the model parameters but nearly $90\%$ of FLOPs, meaning that computation costs are mostly spent at the CONV layers.
As the DWBP overlaps the communication of top layers with the computation of bottom layers,
such structure greatly benefits from the DWBP since $90\%$ of working loads on computation and communication are overlapped, thus the waiting time on GPUs significantly reduces and the GPU utilization greatly increases. We implement the DWBP by creating a separate thread for each independent layer, thereby enable concurrent communications and computations for different layers. The effectiveness of DWBP is empirically evaluated in section \ref{sec:internal_dwbp_sacp}.

\subsection{Structure-Aware Message Passing Protocol}
\label{sec:samp}

%\qirong{What happens when you don't use SF communication? What happens when you use SF communication for ALL messages? Why do we need this hybrid structure-aware protocol?}
%\hao{Both technical details and quantitative analysis are already provided in this section. Add a pointer for the readers to find result in the internal comparison section.}\qirong{Same comment as previous subsection.}
Most ML models, such as neural networks, fall into the family of \emph{matrix-parameterized models} (MPMs), which represent their parameters as a set of matrices.
In data-parallel distributed settings, learning MPMs using iterative-convergent algorithms, as in \cite{Chilimbi:2014:OSDI,Dean:2012:NIPS}, usually needs to repeatedly push out and pull in the whole parameter matrices. Let us take the AlexNet as an example, the weights between the two FC layers \emph{fc6} and \emph{fc7} are represented as a $4,096 \times 4,096$ matrix $W$ as well as its gradients $\nabla W$.　At each iteration, every worker sends out $\nabla W$ and then synchronizes updated $W$, which involves heavily communicating two $4,096 \times 4,096$ float matrices via the network, as Fig.\ref{fig:sf}.(a) shows. However, the commodity Ethernet only affords maximally several Megabits of data being transmitted per second (as in Table \ref{tb:Ethernet}). While in practice, the size of parameters to be communicated grows rapidly with the model size, the problem complexity, and the number of nodes in clusters, and GPU-based computing further deceases the per-iteration computation time. Consequently, the parameters to be transferred per second easily exceed the bandwidth of the network, which in turn bottlenecks the computation. To address this challenge, in Poseidon, besides client-server connections between servers and workers, we also allow P2P connections between every two workers, based on which we design a new communication protocol to minimize the number of parameters needed to be communicated by exploiting a nice property of neural networks.

In this section, we first introduce a novel communication approach of Petuum for distributed machine learning, namely sufficient factor broadcasting (SFB) \cite{Xie:2015:arXiv}, which exchanges parameters following a P2P scheme. Then we discuss the proposed structure-aware message passing protocol, which is essentially a hybrid communication approach between the centralized parameter server (PS) and decentralized SFB. The SCAP significantly minimizes the communication cost by directly reducing the number of parameters needed to be communicated during neural network training, so as to alleviate the bottleneck raised by limited bandwidth of commodity Ethernet. We conduct internal comparisons and demonstrate the effectiveness of SCAP in section \ref{sec:internal_dwbp_sacp}.

\subsubsection{Sufficient Factor-based Communication}
\label{sec:sf}

\begin{figure}[h]
\centering
\includegraphics[width=80mm]{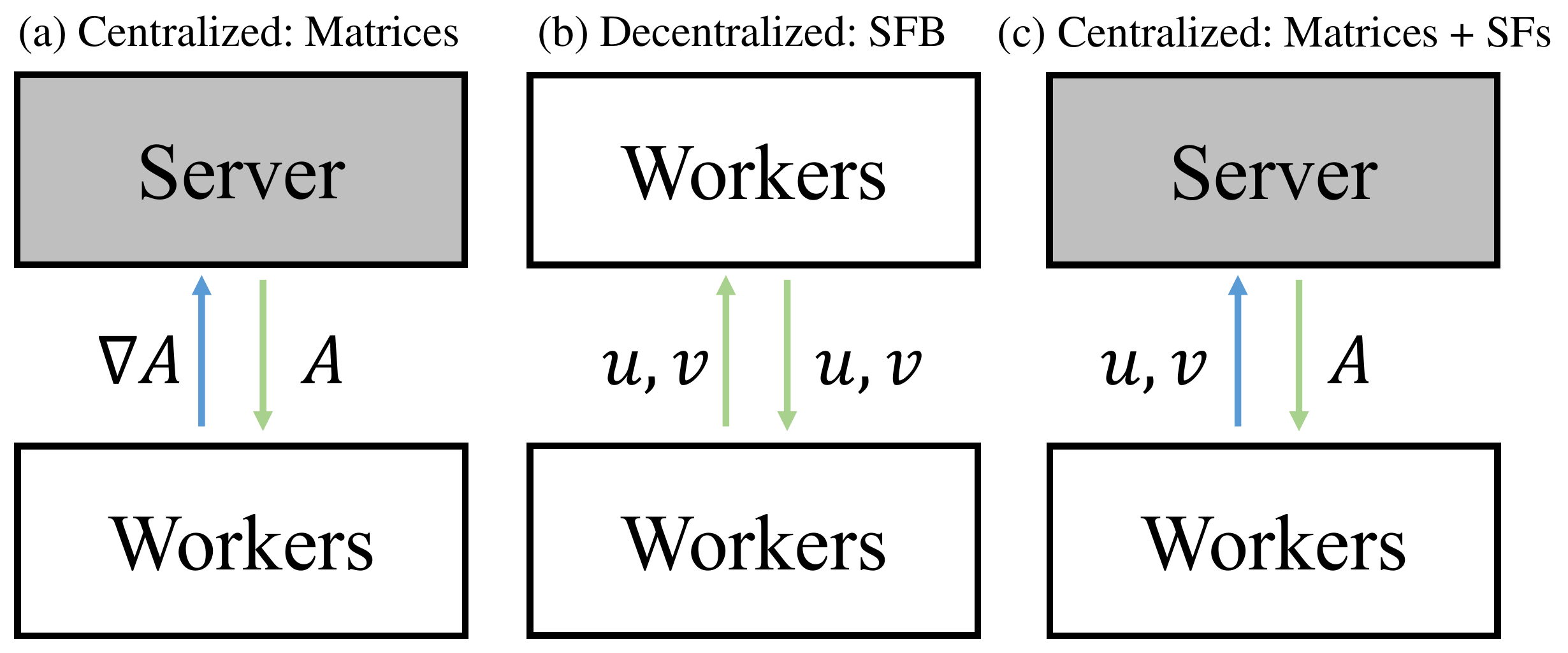}
\caption{\small
The illustration of three types of communications: (a) Full matrices communications via centralized parameter server; (b) Sufficient factor broadcasting via decentralized P2P scheme; (c) SF based communication via centralized parameter server.
}
\label{fig:sf}
\end{figure}

% The bascis of SFB
Some MPMs, including neural networks, enjoy the following structural property:
when training using SGD, their gradient $\nabla W$ over a batch of training samples is a low-rank matrix, which can be casted as the outer product of two vectors $u$ and $v$: $\nabla W = uv^\top$, where $u$ and $v$ are called \textit{sufficient factors} (SFs). Consider the training of CNNs, where $W$ is an $M \times N$ weight matrix between two FC layers $l_i$ and $l_{i+1}$. In the forward pass, one data sample is fed into the network and the activations of layer $l_i$ is produced as $a_i$.
During BP, the loss $\ell$ is propagated, and an error message $E_{i+1}$, which is an $M$ dimensional vector, is passed back from $l_{i+1}$ to $l_i$. The gradients $\nabla W$ thus can be exactly reconstructed by two vectors $E_{i+1}$ and $a_i$:
\begin{equation}
\nabla W = \frac{\partial \ell}{\partial W} = E_{i+1} a_i^\top,
\label{eq:gradient_sf}
\end{equation}
% How SFB system works.
Sufficient factor broadcasting (SFB) ~\cite{Xie:2015:arXiv} is designed to minimize the number of parameters needed to be communicated by leveraging the above property. In a distributed setting with $P$ workers, on worker $p$, instead of directly communicating two full matrices $\nabla W_p$ and $W_p$ with the master node, we recast it to three steps:
(1) Decouple  $\nabla W_p$ into two vectors $u_p$ and $v_p$;
(2) Broadcast $u_p$ and $v_p$ to all other peer workers and also receive sufficient factors $u_i, v_i, i \ne p$ from them, as Fig.\ref{fig:sf}.(b) shows.
(3) Reconstruct $\{\nabla W_i\}_{i=1}^P$ using $\{u_i, v_i\}_{i=1}^P$ as in Eq.(\ref{eq:gradient_sf}), and apply the updates locally on every worker.

% Complexity analysis, in the context of AlexNet
Compared to traditional client-server pipeline, SFB can significantly reduce the communication cost in many popular settings. Consider training a CNN with a batch size of $K$. In each batch, every worker needs to broadcast and receive $K$ sets of $M$ and $N$ dimensional vectors to and from $P-1$ workers, respectively, thus in total $(P-1)^2 K(M+N)$ floats need to be transmitted. While, in a traditional parameter server where the full matrices are sent, the size is $2PMN$ ($P, K \ll M, N$ in modern CNN structures ）. For instance, when training AlexNet on 4 GPU nodes with $K = 256$ and $M, N = 4,096$ for fc6 and fc7, SFB communicates only 18.9M parameters in each iteration, which is $7.1$ times less than communication of full matrices 134.2M.

% Difference from Microsoft Adam
Microsoft Adam \cite{Chilimbi:2014:OSDI} employs a different SF-based strategy. The SFs from all workers are first sent to the master node following the client-server scheme, then transformed into matrices and aggregated to update model parameters. Then, full parameter matrices are sent back to each worker, as Fig.\ref{fig:sf}.(c) shows. Its communication cost is thus $PK(M+N)+ PMN$. With the previous example, 75.5M parameters need to be communicated, which is 4 times larger than SFB.

\begin{figure}[t]
\centering
\hspace{-1em}
\includegraphics[width=82mm]{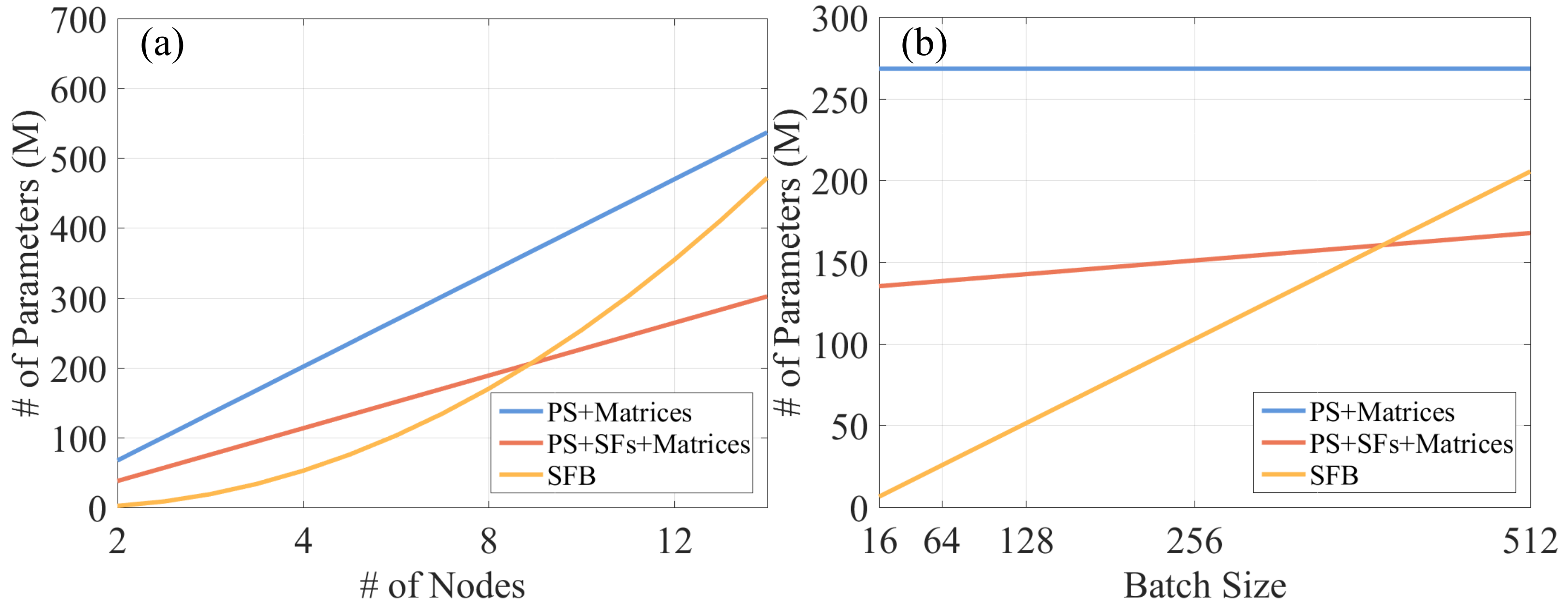}
\caption{\small
Comparisons of the three communication strategies when training AlexNet on GPU clusters. The parameters needed to be communicated between fc6 and fc7 are compared by varying (1) the number of cluster nodes $P$ and (2) batch size $K$.
}
\label{fig:sf_cost}
\end{figure}

Fig.\ref{fig:sf_cost} compares the aforementioned three strategies in terms of the number of parameters needed to be communicated between layer fc6 and fc7 when training Alexnet with different number of nodes and batch size. SFB usually outperforms another two strategies with a smaller batch size. One potential drawback of SFB is that its communication cost increases quadratically with the number of nodes, since it employs the peer-to-peer communication scheme.

\begin{algorithm}[h]
\label{algo:sacp}
\caption{The Structure-aware Communication Protocol (SACP)}
\small
\SetKwInOut{Input}{Input}
\SetKwInOut{Task}{Task}
\setcounter{AlgoLine}{0}
\nonl \hspace{-12pt} \textbf{At iteration $t$ on worker $p$}:\\
\nonl \hspace{-1em}\Input{Layer $l_i$, $M \times N$ gradients $\nabla A_{i}^p$, number of workers $P$, batch size $K$.}
\nonl \hspace{-1em}\Task{Push out gradients $\nabla A_i^p$ and then update $A_i^p$.}
\If{$l_i$ is not an FC layer}{
    Send $\nabla A_i^p$ to the master node.\\
    Synchronize updated $A_i$ from the master node.\\
}\Else{
    Recast $\nabla A_i^p$ into two SFs, \ie $\nabla A_i^p = u_i^p {v_i^p}^\top$;\\
    %Determine the optimal communication approach based on $M$, $N$, $P$ and $K$.
    \If{ $(P-1)^2 K(M+N) \le PK(M + N) + PMN$ }{
        Broadcast $u_i^p, v_i^p$ to all other workers.\\
        Receive SFs $u_i^j, v_i^j, j \ne p$ from all other workers.\\
        Update $A_i$: $A_i \leftarrow A_i + \sum_{j} u_i^j {v_i^j}^\top + \Lambda(A_i)$.\\
    }\Else{
        Send $u_i^p, v_i^p$ to the master node.\\
        Synchronize updated $A_i$ from the master node.\\
    }
}
\end{algorithm}

\subsubsection{Structure-Aware Communication Protocol}
\label{sec:sacp}
We propose the structure-aware communication protocol (SACP), which hybridizes the client-server PS scheme with the P2P SFB scheme, for GPU-based distributed deep learning. The SACP is structure-aware, as it intelligently determines the optimal communication method before communicating the parameters, according to the working layer, the SGD batch size, and the number of workers. In particular, for CONV layers, where layer parameters are sparse, SACP takes the centralized server-client PS scheme to directly communicate the parameters via the parameter server. On the other hand, for FC layers where layer parameters are dense and enjoy the low-rank property of MPMs, the SACP chooses between the two SF-based communication (\ie centralized PS and SFB)  according to the batch size and the number of workers. Algorithm \ref{algo:sacp} summarizes how the SACP intelligently controls the communication.

% supplementary to DWBP
As complementary to the Algorithm \ref{algo:dwbp}, SACP can be synergetically incorporated into DWBP to significantly reduce communication costs as well as improve GPU utilization. Although the SF-based communication may cause extra computation cost due to the reconstruction of gradients from SFs, in GPU based distributed deep learning, such computations are often negligible compared to communication and SF computation.

\subsubsection{Bandwidth Management}
\label{sec:bandwidth}

%\qirong{This should go at the end of the SACP subsection, when you explain that SACP and DWBP are built on top of the bandwidth manager.}
Poseidon also exploits the \textit{B\"{o}sen}-based communication strategy~\cite{Wei:2015:SoCC}, a key component of Petuum that maximizes the network efficiency under a given network bandwidth budget (especially in commodity Ethernet ) while minimizing parallel errors.
Cooperating with DWBP and SACP, which are aware of the model and cluster structures, the bandwidth manager further incorporates the prior knowledge on the low-level network bandwidth, and maximizes communication efficiency by prioritizing network bandwidth for messages most significant for algorithm progress. Specifically, it communicates model updates and dirty model parameters as quickly as possible without overusing the network bandwidth budget (full network utilization), and allocates network bandwidth according to the messages' contribution to convergence. In Poseidon, the bandwidth manager lies at the bottom of DWBP and SACP (as shown in Figure \ref{fig:overview}), and manages the message passing among server and clients regardless of the message types (matrices or SFs).

\subsection{Other Features}
\label{sec:other}
%\qirong{The tone here makes it sound like you are just squeezing in text for no good reason. See my edits.}
%st{We briefly present some additional ingredients and utilities of Poseidon. Since some of them are slightly out of focus of this paper, we skip the details that can be found in}
Poseidon includes features to enhance the usability of the deep learning software system, by addressing issues such as distributed storage and fault tolerance. While not crucial to the performance of distributed GPU-based training, they help to improve the user experience.

\noindent \textbf{Distributed Storage}.
%\qirong{No need for ``utilities" heading once you move bandwidth management to section 3.}
Poseidon allows both shared and private file systems for multiple cluster nodes, so that the training data can be stored either in a shared file system to  be simultaneously accessed by all cluster nodes, or in separate file systems that each node has a separate data partitions, to avoid I/O overload.

\noindent \textbf{Fault Tolerance}. Poseidon provides fault tolerance by checkpointing all clients' model states. Either in the event of failure or as the user specifies, the entire distributed CNN system can be restarted from the last checkpoint exactly, keeping all model/solver states and database pointers unchanged as before.

\section{Evaluation}
\label{sec:evaluation}
We first evaluate Poseidon on image classification tasks with benchmark datasets of CIFAR-10~\cite{Krizhevsky:2009:cifar} and ILSVRC2012~\cite{Russakovsky:2015:IJCV},
and show that Poseidon significantly accelerates the training of modern CNN structures, while guaranteeing the correct convergence, which is important for distributed deep learning.
Moreover, we deploy Poseidon on the ImageNet 22K classification, and compare its performance with
%those of \textit{MXNet}~\cite{MXNet}
previously published results such as Adam~\cite{Chilimbi:2014:OSDI}.
Finally, we conduct some internal comparisons to justify the effectiveness of DWBP and SACP.

\noindent \textbf{Cluster Configuration}.
We conduct all experiments on the PRObE Susitna cluster~\cite{Lloyd:2013:NSDI}, where each node has $4\times 16$-core 2.1GHz AMD Opteron 6272 CPUs, 128GB of RAM, and NVIDIA Tesla K20C GPU with 4799MB memory. All cluster nodes have shared access to a NFS with 1x Hitachi 1.0 TB HDD and 2x Hitachi 3.0 TB HDD.
We use the 40GbE network for both connecting NFS and communication among workers.
For software, we use the Caffe version at Oct 2014 with CUDA 6.5 and CUDNN R2.
and NVIDIA driver version 340.29.

\subsection{Image Classification}
We demonstrate Poseidon's performance on three benchmark datasets ranging from small to large, including the CIFAR-10 \cite{Krizhevsky:2009:cifar}, the ILSVRC2012 and the ImageNet22K\cite{Russakovsky:2015:IJCV}. The statistics of the datasets are briefly summarized in Table \ref{tb:dataset_stat}.

\subsubsection{Classification on CIFAR-10}
\label{sec:class_cifar}

We first evaluate our Poseidon on the CIFAR-10 dataset, which contains $32 \times 32$ images of $10$ classes, with 6K images per class.
An official train/test split is provided that 50K images are used for training and 10K for testing.
Although CIFAR-10 is a relatively small dataset, we experiment to show Poseidon's capability on achieving better accuracy than single machine at the same time accelerate the training of small CNNs.

\noindent \textbf{Settings}.
We employ the built-in \textit{cifar10\_quick\_solver} and \textit{cifar10\_quick\_train\_test} network structure in Caffe\footnote{{\scriptsize\url{github.com/BVLC/caffe/tree/master/examples/cifar10}}.}, consisting of 3 CONV layers and 1 FC layers followed by a 10-way softmax classifier, in total $145,578$ parameters.
It converges to a $70\%$ test accuracy with 4 epochs of training in a single machine without decreasing the learning rate.
We deploy Poseidon onto 8 Susitna nodes. As a larger batch size usually hurts the SGD performance, for both settings, we reduce the batch size from 100 to 50 and also slightly decrease the base learning rate from 0.01 to 0.007, while keeping other solver settings unchanged. All CIFAR-10 images are stored in a single LMDB on NFS with shared access to 8 nodes. For better comparison, in the distributed setting, we set the staleness $s$ to zero
(\ie we use BSP consistency model during training).

\noindent \textbf{Performance}.
Similar to the single machine setting, we train the network to convergence without adjusting the learning rate. The test accuracy achieves nearly $75\%$. Figure.\ref{fig:convergence}(a)-(b) plots how the test error decreases along with training time and iterations for Poseidon on 8 nodes and Caffe on a single node.
Under the same setting, the single machine Caffe takes more than $4$ times of training time to converge to $70\%$ accuracy, while Poseidon quickly converges to $72\%$ in 19 seconds and attain a higher accuracy $75\%$ in $25$ seconds with $8$ GPU nodes.

\begin{table}[t]
\footnotesize
\centering
	\begin{tabular}{|c|c|c|c|}
	\hline
	Dataset & \# of Images  & Size of images & \# of categories\\ \hline \hline
	{\textbf{CIFAR-10} } & 60K  &  $32 \times 32 \times 3$ & 10 \\ \hline
	{\textbf{ILSVRC2012} } & 1.3M & $256 \times 256 \times 3$ & 1000\\ \hline
  	{\textbf{ImageNet22K} } & 14.2M & $256 \times 256 \times 3$ & 21841\\ \hline
    \end{tabular}
\caption{Statistics of the benchmark datasets we use for evaluation of the performance of Poseidon.}
\label{tb:dataset_stat}
\end{table}

\begin{table}[t]
\footnotesize
\centering
	\begin{tabular}{|c|c|c|c|c|}
	\hline
	Model  & Speedup & top-1 accuracy\\ \hline \hline
	\textbf{CIFAR-10 quick} & $4 \times$  & $75 \%$ \\ \hline
	\textbf{AlexNet}   & $4.5 \times$  & $56.5\%$ \\ \hline
  	\textbf{GoogLeNet} & $4 \times$ &  $67.1\%$ \\ \hline
    \end{tabular}
\vspace{3pt}
\caption{Speedups and converged top-1 accuracies of Poseidon by training models on 8 GPU nodes with CIFAR-10 and ILSVRC 2012 dataset, compared to Caffe on a single machine.}
\label{tb:results}
\end{table}

\subsubsection{Classification on ILSVRC 2012}
\label{sec:class_ilsvrc}

We then experiment on ImageNet ILSVRC 2012, consisting of 1.28 million of training images and 50K validation images over 1,000 categories.
Following the standards, we downsample all images to $256 \times 256 \times 3$ before feeding into the networks, and report the top-1 accuracy on the validation set. These experiments show that Poseidon significantly accelerates the training of modern state-of-the-art CNN architectures at the same time guarantees the correct convergence in a distributed GPU cluster.

\noindent \textbf{Settings}.
We evaluate Poseidon using AlexNet \cite{Krizhevsky:2012:NIPS} and GoogLeNet \cite{Szegedy:2014:going}. The AlexNet is a de facto standard CNN architecture with 5 CONV layers, 2 FC layers and a 1000-class softmax classifier, in total 61.3 million of parameters. GoogLeNet is a more structural and deeper (22-layer) CNN with only 5 million of parameters by stacking inception modules \cite{Szegedy:2014:going}. For fair comparisons, we employ the open implementations of AlexNet\footnote{\scriptsize \url{github.com/BVLC/caffe/tree/master/models/bvlc_alexnet}.} and GoogLeNet\footnote{\scriptsize\url{github.com/BVLC/caffe/tree/master/models/bvlc_googlenet}.} provided in Caffe. Specifically, the \textit{bvlc\_alexnet} achieves $57\%$ top-1 accuracy after convergence, and the \textit{bvlc\_googlenet} converges to $68.7\%$ top-1 accuracy, both using just the center crop for testing. In single machine training, for AlexNet, we use the standard solver in Caffe, which trains with a batch size 256 for nearly 70 epochs, during which the learning rate is decreased by dividing 10 for 3 times. For GoogLeNet, we employ the \textit{quick\_solver}, which uses the polynomial learning rate policy, and trains for 60 epochs with batch size set to $32$. In the distributed setting, we deploy both AlexNet and GoogLeNet onto 8 GPU nodes with fully data-parallel training, and keep the network structure and the batch size exactly the same, but change to a more suitable solver setting. Specifically, for AlexNet, we train on 8 nodes for about 60K iterations, with the base learning rate set to 0.005 and decreased 5 times during the whole training. For GoogLeNet, we use a standard step policy by setting the base learning rate to $0.005$ and decrease $90$ times during training. Using a single LMDB on NFS bottlenecks training when it is simultaneously read by 8 nodes, thereby we split it into 8 parts and let every node access a separate part to avoid I/O overload.

\begin{figure}[h]
%\hspace{-3em}
\includegraphics[width=80mm]{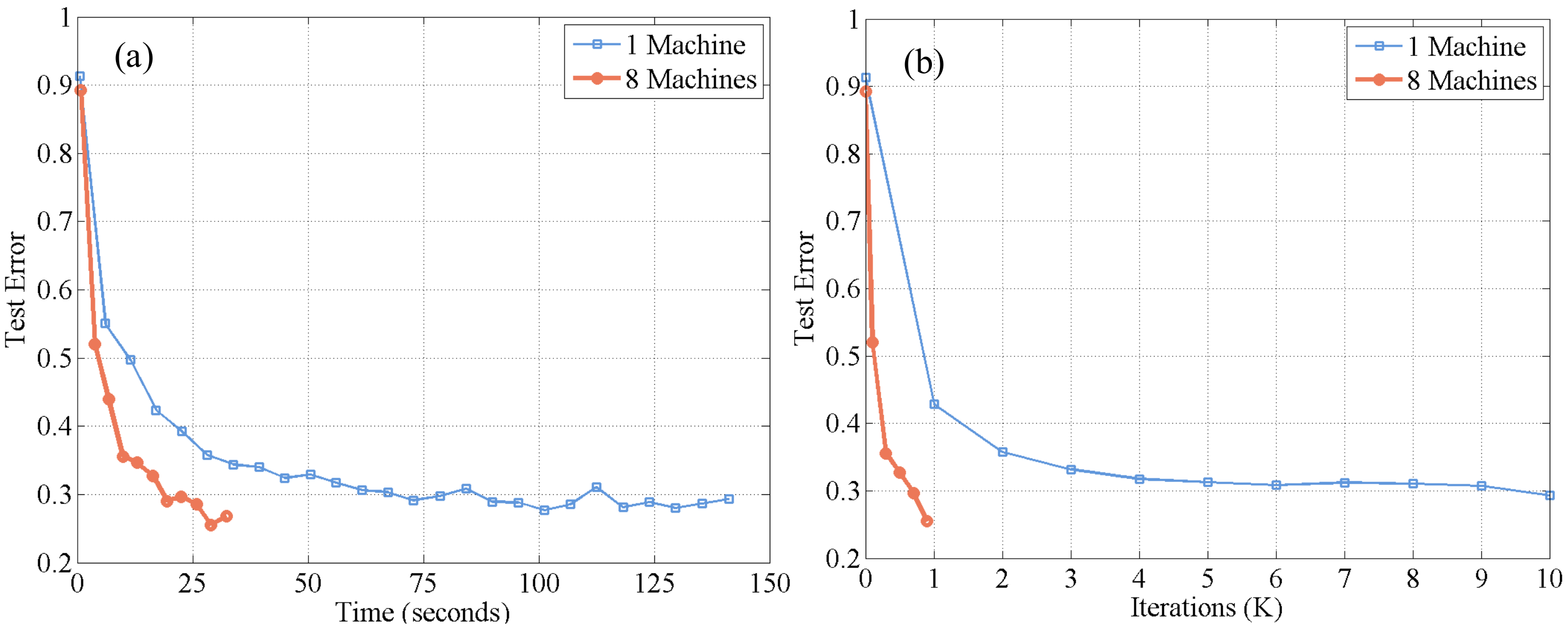}
\includegraphics[width=80mm]{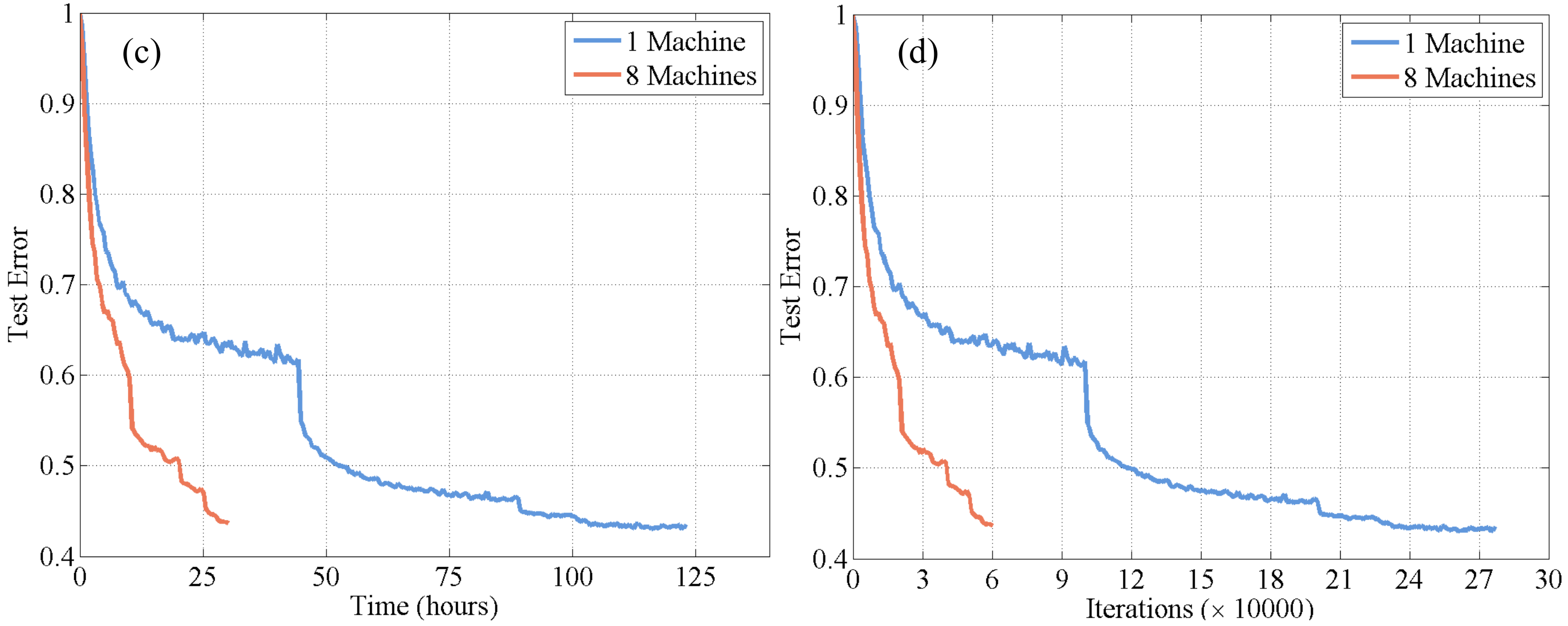}
\includegraphics[width=80mm]{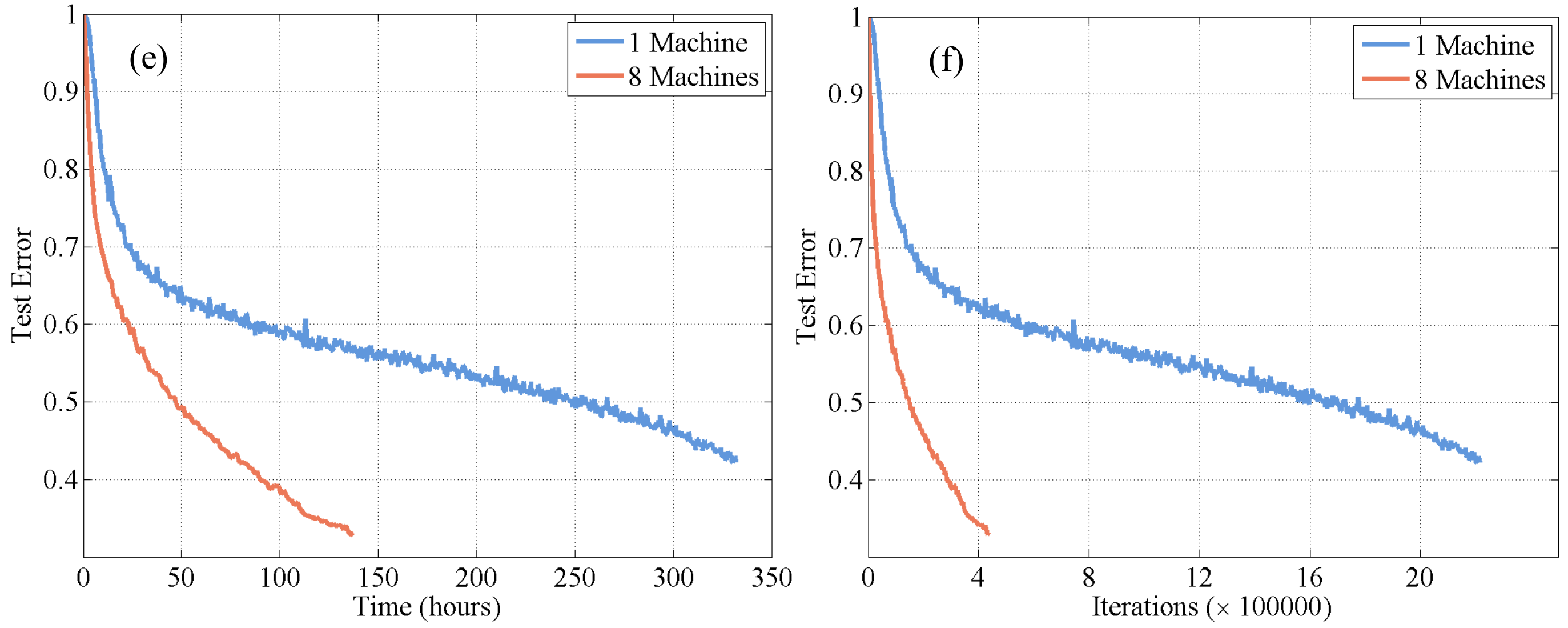}
\caption{
The comparison of training different CNNs to convergence between Poseidon on 8 GPU nodes and Caffe on a single node: (a)-(b) \textit{cifar10\_quick\_train\_test}; (c)-(d) \textit{bvlc\_alexnet}; (e)-(f) \textit{bvlc\_googlenet}.
Test errors are shown with respect to (left) training time, and  (right) training iterations.
}
\label{fig:convergence}
\end{figure}

\noindent \textbf{Performance}.
Figure.\ref{fig:convergence}(c)-(d) and Figure.\ref{fig:convergence}(e)-(f) show the performance of training AlexNet and GoogLeNet using Poseidon with a GPU cluster of 8 nodes, compared to single machine Caffe, respectively. For AlexNet, Poseidon attains $56.5\%$ top-1 accuracy on the validation set after training of 27 hours, with a $4.5 \times$ speedup as compared to single machine Caffe that needs 5 days. For GoogLeNet, Poseidon converges to $67.1\%$ top-1 accuracy after 130 hours of training, as compared to Caffe, which only achieves $50\%$ top-1 accuracy after 250 hours of training, and $57\%$ after near 350 hours of training on a single Susitna node (Poseidon only needs less than 48 hours to achieve $50\%$ and 75 hours to achieve $57\%$, with a near $5 \times$ speedup), and hard to converge with more than 500 hours of training.
Finally, we summarize the convergence speedups of Poseidon in Table.\ref{tb:results}.

\subsubsection{Classification on ImageNet 22K}
\label{sec:class_imagenet22k}
ImageNet 22K is the largest public dataset for image classification, including 14,197,087 labeled images from 21,841 categories, which is rarely touched by the research community due to its massive data size and complexity.
We experiment on ImageNet 22K to demonstrate the scalability of Poseidon. As no official test data exists for evaluation, following previous settings in \cite{Chilimbi:2014:OSDI,Dean:2012:NIPS,Le:2012:ICML}, we randomly split the whole set into two parts, and use the first 7.1 million of images for training and remained for test. Similar to ILSVRC 2012, we resize all images to $256 \times 256$  and report the top-1 test accuracy.
%Note that random guess from 21,841 categories can only achieve $0.0045\%$ accuracy on this task.

\noindent\textbf{Settings}.
We design a AlexNet-like CNN architecture; specifically, the CNN takes a random $227 \times 227 \time 3$ crop from the original image, and forwards it into 5 CONV layers and 2 FC layers before making a prediction.
The CNN has convolution filters with sizes $7 \times 7, 5 \times 5$ and $3 \times 3$. Similar to AlexNet, the first, second and fifth CONV layers are followed by max pooling layers with size $3 \times 3$ and stride 2.
Two FC layers with 3,000 neurons each are put at the top of the network, followed by a softmax layer to be a 21,841-way classier with 120M parameters and 1.8 billion of connections overall.
We train the CNN with fully data-parallelism by equally partitioning and distributing the training data into 8 GPU nodes. The batch size and staleness are fixed at 256 and 0, respectively. The network is trained using the step learning rate policy, with base learning rate set to 0.005 and decreased 6 times.

\noindent \textbf{Performance}.
Table \ref{tb:imagenet22k} compares our result to those of previous work on ImageNet 22K, Adam \cite{Chilimbi:2014:OSDI}, MXNet, and Le \etal\cite{Le:2012:ICML}.
Note that at this point complete fair comparison between different framework is not possible,
because the experiment protocol of ImageNet 22K is not standardized, all the source codes are not fully available yet, and large variations exist in system configurations, models, and implementation details.
However, it is clear that Poseidon achieves a competitive accuracy $23.7\%$ with the state-of-the-arts with shorter training time and less machine resources. Compared to Adam \cite{Chilimbi:2014:OSDI}, we only use $30\%$ training time and $13\%$ machines to achieve $23.7\%$ accuracy with a similar sized model.
%although train accuracy may not be a correct benchmark index in image classification tasks,
Promisingly, we achieve a higher training accuracy with 3 days of training using a well-established CNN model --- this which compares favorably to MXNet, which uses the whole set of 14.1 million images to train an inception-BN structure \cite{ioffe2015batch} using 4 GPUs in a single machine without network communication, and reports $37.1\%$ train accuracy after 8.5 days of training.

\begin{table*}[t]
\footnotesize
\centering
	\begin{tabular}{|K{1.8cm}|K{5.5cm}|K{2cm}|c|c|c|c|c|c|}
	\hline
	Framework & Data  & \# machines/cores & Time & Train accuracy & Test accuracy \\ \hline \hline
	\textbf{Poseidon} &  7.1M ImageNet22K for training, 7.1M for test  & 8 / 8 GPUs  & 3 days & $41\%$ & $23.7\%$ \\ \hline
	\textbf{Adam \cite{Chilimbi:2014:OSDI}} & 7.1M ImageNet22K for training, 7.1M for test & 62 machines/?  &  10 days & N/A & $29.8\%$ \\ \hline
    \textbf{MxNet \cite{MXNet}} & All ImageNet22K images for training, no test & 1/4 GPUs & 8.5 days & $37.19 \%$ & N/A \\ \hline
    \textbf{Le \etal \cite{Le:2012:ICML}} \textbf{w/ pretrain} & 7.1M ImageNet 22K, 10M unlabeled images for training, 7.1M for test & 1,000/1,6000 CPU cores &  3 days & N/A & $15.8\%$ \\ \hline
    \end{tabular}
\vspace{3pt}
\caption{Comparisons of the image classification results on ImageNet 22K.}
\vspace{-9pt}
\label{tb:imagenet22k}
\end{table*}

\subsection{Internal Comparisons}
\begin{figure}[h]
\includegraphics[width=80mm]{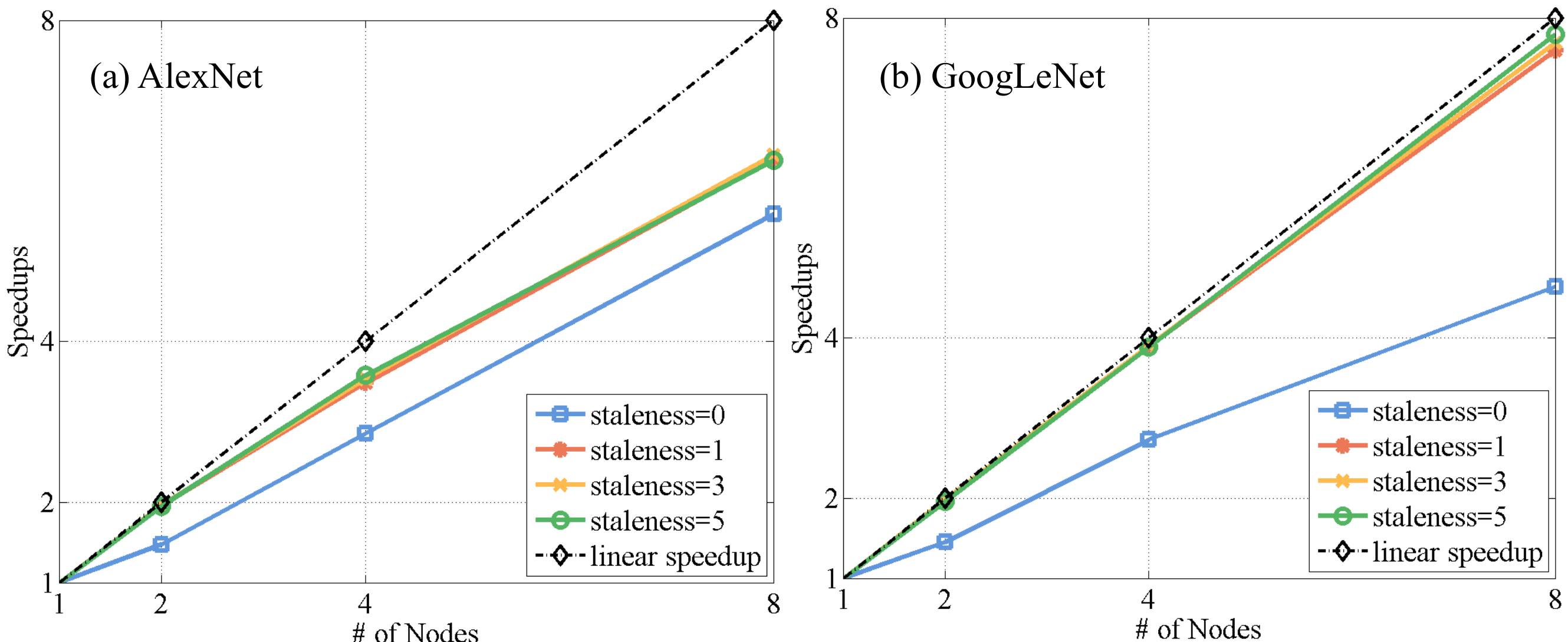}
\caption{
The speedups on throughput with different values of staleness, when training using Poseidon on 8 nodes, compared to Caffe on a single node.
(a) AlexNet with batch size 256, and  (b) GoogLeNet with batch size 32.
}
\label{fig:throughput}
\end{figure}

In this section, we conduct internal comparisons to study the effectiveness of DWBP and SACP in improving the GPU utilization, as well as reducing communication cost for GPU-based distributed deep learning. Besides, we report the speedups on throughput (\ie number of images trained per seconds) in Fig.\ref{fig:throughput} when training AlexNet and GoogLeNet using Poseidon on 8 GPU nodes with different staleness settings, compared to single machine Caffe.

\subsubsection{DWBP and SACP}
\label{sec:internal_dwbp_sacp}

Since DWBP executes asynchronously in a multi-thread and multi-machine setting, it's difficult to directly monitor how the communication and computation are overlapped. To measure the improvement by DWBP and SACP, we instead evaluate the speedups on throughput, which is defined as the number of images processed per second given a model and a batch size, compared to the single machine Caffe.

\begin{figure}[t]
\includegraphics[width=80mm]{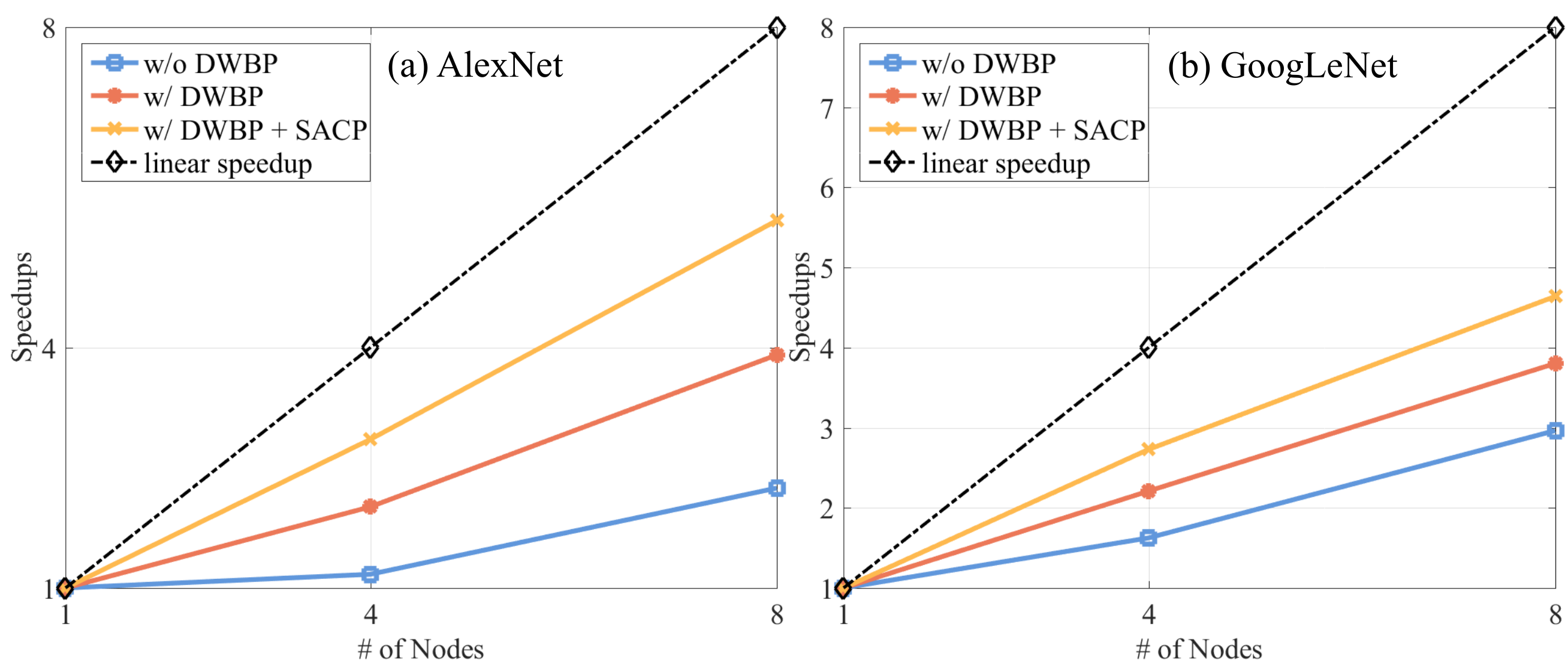}
\caption{
The internal comparisons of training AlexNet and GoogLeNet using Poseidon with different number of GPU nodes and settings: (a) AlexNet with batch size $256$ ; (b) GoogLeNet with batch size $32$.  When running on only $1$ node, the performance of original Caffe is reported.
}
\label{fig:internal}
\end{figure}

Fig.\ref{fig:internal} compares the speedups for training AlexNet and GoogLeNet under the following three settings with different number of nodes: (1) {\small \tt w/o DWBP}: parallel training with traditional BP and full matrices communication; (2) {\small \tt w/ DWBP}: parallel training with DWBP enabled; (3) {\small \tt w/ DWBP + SACP}, parallel training with both DWBP and SACP enabled. We follow the standard setting, \ie we set the staleness to $0$ (BSP), and the batch size to $256$ for AlexNet and $32$ for GoogLeNet \footnote{Different batch sizes will lead to slightly different speedups on throughput.}. Obviously, with DWBP to overlap the communication with computation, the waiting time between two iterations is greatly saved, thus the throughput is significantly improved, thereby the GPU utilization ratio is relatively improved. Specifically, as Fig.\ref{fig:internal}.(a) shows, for AlexNet, when training using 8 nodes, DWBP significantly improve the speedup from $2.2$ to $4$, with nearly $2 \times$ more speedups. For GoogLeNet with less parameters, DWBP also brings $30 \%$ more speedups.

With SACP enabled, the speedup on throughput is further improved. Particularly, when training on 8 nodes, although SACP may bring extra computation costs due to parameter matrix reconstructions, it still greatly increases the speedups of AlexNet training from $4$ to $6$, with a $50\%$ improvement. For GoogLeNet with fewer FC layers, SACP provides approximately $20\%$ improvement on the speedup.

It is clear to see that, we will suffer more loss on the throughput when increasing the number of nodes. Specifically, when directly parallelizing AlexNet on a 8-node GPU cluster without any system/algorithm optimization, we suffer a $80\%$ loss in throughput, comparing to the ideally linear speedup. However, with DWBP and SACP enabled, we only suffer less than $25\%$ loss, which makes Poseidon much closer to the linear speedup.

\subsubsection{SSP Consistency Model}
In this section, we study the efficacy of stale synchronous parallel (SSP) consistency model, which is a unique feature provided by Petuum, on scaling up distributed deep learning. Specifically, we compare the speedup on throughput of training AlexNet and GoogLeNet using Poseidon by varying the value of the staleness threshold $s$, while keeping all other settings　fixed. Setting staleness values to zero (\ie $s = 0$) leads the consistency management to be \textit{bulk synchronous parallelization} (BSP), where computation uses local model copies that are synchronized only at the end of each iteration and the next iteration may not start  until all machines have received up-to-date model parameters. Therefore, with BSP the learning speed is limited by the slowest machine.
Compared to the BSP, a positive staleness value produces a short grace period for parameter synchronization between every two iterations, thus enables us to manage the bandwidth for parameter exchanges according to current bandwidth budget and the dirtiness of the updates.

As seen in Fig.\ref{fig:throughput}.(a), for AlexNet training where communication load is quite heavy, if we set a positive value of $s$, the throughput is greatly improved; with 4 nodes, the speedup of the fully BSP ($s = 0$) is improved from $3$ to $3.8$ ($s = 1$). For GoogLeNet training on Poseidon in Fig.\ref{fig:throughput}.(b), a positive value of $s$ makes Poseidon agnostic to communication cost \ie we can enjoy near linear speedups of throughput.

\vspace{-6pt}
\section{Conclusion}
\label{sec:conclusion}
We present Poseidon, a highly scalable and efficient system architecture for large-scale deep learning on GPU clusters. Poseidon is built upon Petuum, thus inherits many functionaries and benefits of Petuum. Its design focuses on efficiently harnessing multiple, distributed GPUs on commodity hardware and Ethernet, in order to maximally scale up existing single-machine DL frameworks  with a fully data parallel scheme for distributed deep learning.
We empirically evaluate Poseidon regarding of throughput, convergence and accuracy on the image classification tasks with multiple standard datasets, and show that Poseidon is able to achieve state-of-the-art speedups in accelerating the training of modern CNN structures, at the same time guarantee the correct convergence.

\begin{comment}
\section{Acknowledgments}
This work is supported by NSF Award XXX.
\end{comment}

\clearpage \newpage
{\footnotesize \bibliographystyle{acm}
\bibliography{poseidon}}

\begin{thebibliography}{10}

\bibitem{Bergstra:2011:NIPSW}
{\sc Bergstra, J., Bastien, F., Breuleux, O., Lamblin, P., Pascanu, R.,
  Delalleau, O., Desjardins, G., Warde-Farley, D., Goodfellow, I.~J., Bergeron,
  A., and Bengio, Y.}
\newblock {Theano: Deep Learning on GPUs with Python}.
\newblock In {\em NIPSW\/} (2011).

\bibitem{Chilimbi:2014:OSDI}
{\sc Chilimbi, T., Apacible, Y. S.~J., and Kalyanaraman, K.}
\newblock {Project Adam: Building an Efficient and Scalable Deep Learning
  Training System}.
\newblock In {\em OSDI\/} (2014).

\bibitem{Coates:2013:ICML}
{\sc Coates, A., Huval, B., Wang, T., Wu, D.~J., Ng, A.~Y., and Catanzaro, B.}
\newblock {Deep Learning with COTS HPC Systems}.
\newblock In {\em ICML\/} (2013).

\bibitem{Collobert:2011:NIPSW}
{\sc Collobert, R., Kavukcuoglu, K., and Farabet, C.}
\newblock {Torch7: A Matlab-like Environment for Machine Learning}.
\newblock In {\em NIPSW\/} (2011).

\bibitem{Dai:2015:Analysis}
{\sc Dai, W., Kumar, A., Wei, J., Ho, Q., Gibson, G., and Xing, E.~P.}
\newblock Analysis of high-performance distributed ml at scale through
  parameter server consistency models.
\newblock In {\em Proceedings of the 29th AAAI Conference on Artificial
  Intelligence\/} (2015).

\bibitem{Dean:2012:NIPS}
{\sc Dean, J., Corrado, G.~S., Monga, R., Chen, K., Devin, M., Le, Q.~V., Mao,
  M.~Z., Ranzato, M., Senior, A., Tucker, P., Yang, K., and Ng, A.~Y.}
\newblock {Large Scale Distributed Deep Networks}.
\newblock In {\em NIPS\/} (2012).

\bibitem{Deng:2013:ICASSP}
{\sc Deng, L., Li, J., Huang, J.-T., Yao, K., Yu, D., Seide, F., Seltzer,
  M.~L., Zweig, G., He, X., Williams, J., Gong, Y., and Acero, A.}
\newblock {Recent Advances in Deep Learning for Speech Research at Microsoft}.
\newblock In {\em ICASSP\/} (2013).

\bibitem{fbcunn}
{\sc {Facebook AI Research}}.
\newblock \url{https://github.com/facebook/fbcunn}.

\bibitem{Ho:2013:NIPS}
{\sc Ho, Q., Cipar, J., Cui, H., Kim, J.~K., Lee, S., Gibbons, P.~B., Gibson,
  G.~A., Ganger, G.~R., and Xing, E.~P.}
\newblock {More Effective Distributed ML via a Stale Synchronous Parallel
  Parameter Server}.
\newblock In {\em NIPS\/} (2013).

\bibitem{ioffe2015batch}
{\sc Ioffe, S., and Szegedy, C.}
\newblock Batch normalization: Accelerating deep network training by reducing
  internal covariate shift.
\newblock {\em arXiv preprint arXiv:1502.03167\/} (2015).

\bibitem{Jia:2014:MM}
{\sc Jia, Y., Shelhamer, E., Donahue, J., Karayev, S., Long, J., Girshick, R.,
  Guadarrama, S., and Darrell, T.}
\newblock {Caffe: Convolutional Architecture for Fast Feature Embedding}.
\newblock In {\em MM\/} (2014).

\bibitem{Krizhevsky:2009:cifar}
{\sc Krizhevsky, A.}
\newblock {Learning Multiple Layers of Features from Tiny Images}.
\newblock Master's thesis, University of Toronto, 2009.

\bibitem{Krizhevsky:2014:arXiv}
{\sc Krizhevsky, A.}
\newblock {One Weird Trick for Parallelizing Convolutional Neural Networks}.
\newblock In {\em arXiv:1404.5997\/} (2014).

\bibitem{Krizhevsky:2012:NIPS}
{\sc Krizhevsky, A., Sutskever, I., and Hinton, G.~E.}
\newblock {ImageNet Classification with Deep Convolutional Neural Networks}.
\newblock In {\em NIPS\/} (2012).

\bibitem{Kumar:2014:Fugue}
{\sc Kumar, A., Beutel, A., Ho, Q., and Xing, E.~P.}
\newblock Fugue: Slow-worker-agnostic distributed learning for big models on
  big data.

\bibitem{Le:2012:ICML}
{\sc Le, Q.~V., Monga, R., Devin, M., Chen, K., Corrado, G.~S., Dean, J., and
  Ng, A.~Y.}
\newblock {Building High-level Features Using Large Scale Unsupervised
  Learning}.
\newblock In {\em ICML\/} (2012).

\bibitem{Lloyd:2013:NSDI}
{\sc Lloyd, W., Freedman, M.~J., Kaminsky, M., and Andersen, D.~G.}
\newblock {Stronger Semantics for Low-Latency Geo-Replicated Storage}.
\newblock In {\em NSDI\/} (2013).

\bibitem{Mikolov:2013:ICLRW}
{\sc Mikolov, T., Chen, K., Corrado, G., and Dean, J.}
\newblock {Efficient Estimation of Word Representations in Vector Space}.
\newblock In {\em ICLRW\/} (2013).

\bibitem{moritz2015sparknet}
{\sc Moritz, P., Nishihara, R., Stoica, I., and Jordan, M.~I.}
\newblock Sparknet: Training deep networks in spark.
\newblock {\em arXiv preprint arXiv:1511.06051\/} (2015).

\bibitem{MXNet}
{\sc MXNet}.
\newblock \url{http://mxnet.readthedocs.org/}.

\bibitem{rumelhart1985learning}
{\sc Rumelhart, D.~E., Hinton, G.~E., and Williams, R.~J.}
\newblock Learning internal representations by error propagation.
\newblock Tech. rep., DTIC Document, 1985.

\bibitem{Russakovsky:2015:IJCV}
{\sc Russakovsky, O., Deng, J., Su, H., Krause, J., Satheesh, S., Ma, S.,
  Huang, Z., Karpathy, A., Khosla, A., Bernstein, M., Berg, A.~C., and Fei-Fei,
  L.}
\newblock {ImageNet Large Scale Visual Recognition Challenge}.
\newblock {\em IJCV\/} (2015), 1--42.

\bibitem{Simonyan:2015:ICLR}
{\sc Simonyan, K., and Zisserman, A.}
\newblock {Very Deep Convolutional Networks for Large-Scale Image Recognition}.
\newblock In {\em ICLR\/} (2015).

\bibitem{Szegedy:2014:going}
{\sc Szegedy, C., Liu, W., Jia, Y., Sermanet, P., Reed, S., Anguelov, D.,
  Erhan, D., Vanhoucke, V., and Rabinovich, A.}
\newblock Going deeper with convolutions.
\newblock In {\em CVPR\/} (2015).

\bibitem{Vasilache:2015:ICLR}
{\sc Vasilache, N., Johnson, J., Chintala, S., Piantino, S., and LeCun, Y.}
\newblock {Fast Convolutional Nets With fbfft: A GPU Performance Evaluation}.
\newblock In {\em ICLR\/} (2015).

\bibitem{Wei:2015:SoCC}
{\sc Wei, J., Dai, W., Qiao, A., Ho, Q., Cui, H., Ganger, G.~R., Gibbons,
  P.~B., Gibson, G.~A., and Xing, E.~P.}
\newblock {Managed Communication and Consistency for Fast Data-parallel
  Iterative Analytics}.
\newblock In {\em SoCC\/} (2015).

\bibitem{Xie:2015:arXiv}
{\sc Xie, P., Kim, J.~K., Zhou, Y., Ho, Q., Kumar, A., Yu, Y., and Xing, E.}
\newblock {Distributed Machine Learning via Sufficient Factor Broadcasting}.
\newblock In {\em arXiv\/} (2015).

\bibitem{Xing:2015:KDD}
{\sc Xing, E.~P., Ho, Q., Dai, W., Kim, J.~K., Wei, J., Lee, S., Zheng, X.,
  Xie, P., Kumar, A., and Yu, Y.}
\newblock {Petuum: A New Platform for Distributed Machine Learning on Big
  Data}.
\newblock In {\em KDD\/} (2015).

\bibitem{Yadan:2014:ICLRW}
{\sc Yadan, O., Adams, K., Taigman, Y., and Ranzato, M.}
\newblock {Multi-GPU Training of ConvNets}.
\newblock In {\em ICLRW\/} (2014).

\bibitem{Zou:2014:VLDB}
{\sc Zou, Y., Jin, X., Li, Y., Guo, Z., Wang, E., and Xiao, B.}
\newblock {Mariana: Tencent Deep Learning Platform and its Applications}.
\newblock In {\em VLDB Endowment\/} (2014).

\end{thebibliography}

%\bibliographystyle{abbrvnat}
%\bibliography{Poseidon}
% The bibliography should be embedded for final submission.
%\theendnotes

\end{document}